%% file: main.tex
\documentclass[conference]{IEEEtran}
\usepackage{times}
\usepackage{subcaption}

\usepackage[numbers,sort]{natbib}
\usepackage{multicol}
\usepackage[bookmarks=true]{hyperref}
\usepackage{amsmath}
\usepackage[capitalise]{cleveref}
\usepackage{xcolor}
\usepackage{graphicx}
\usepackage{booktabs}
\usepackage{float}
\usepackage{enumerate}
\usepackage[most]{tcolorbox}
\usepackage{lipsum}
\usepackage{adjustbox}
\usepackage{subcaption}
\usepackage{listings}
\usepackage{siunitx}
\usepackage{pifont}
\usepackage[utf8]{inputenc} 
\usepackage{amssymb}
\usepackage{bm}
\usepackage{cuted}
\usepackage{tikz}
\usepackage{glossaries}

\usepackage{multirow}
\usepackage{makecell}
\newcommand{\tabincell}[2]{\begin{tabular}{@{}#1@{}}#2\end{tabular}}

\input{acr.tex}
\input{bubbles.tex}

\newcommand{\cmark}{\ding{51}}%
\newcommand{\xmark}{\ding{55}} 

\usepackage{eso-pic}

\newcommand\AtPageUpperCenterNotice[1]{%
  \AtPageUpperLeft{%
    \put(\LenToUnit{0.5\paperwidth},\LenToUnit{-2cm}){\makebox[0pt]{#1}}%
  }%
}

\AddToShipoutPictureBG*{%
  \AtPageUpperCenterNotice{%
    \parbox[b][2cm][c]{\paperwidth}{%
      \centering
      \fontsize{12}{14}\selectfont
      \color{gray!50}
      This paper has been accepted for publication at the\\
      Robotics Science and Systems (RSS), Los Angeles 2025. 
    }%
  }%
}

\begin{document}

\title{Enhancing Autonomous Driving Systems with On-Board Deployed Large Language Models \vspace{-0.5cm}}

\author{
  Nicolas Baumann$^{*, \dagger}$,
  Cheng Hu$^{\ddagger}$,
  Paviththiren Sivasothilingam$^{*}$,
  Haotong Qin$^{*}$,
  Lei Xie$^{\ddagger}$,
  Michele Magno$^{*}$,
  Luca Benini$^{\dagger}$
  \\
  $^{*}$Center for Project-Based Learning, ETH Zurich, Switzerland \\
  $^{\dagger}$Integrated Systems Laboratory, ETH Zurich, Switzerland \\
  $^{\ddagger}$Department of Control Science and Engineering, Zhejiang University, China \\
}


%

\maketitle

\begin{strip}
\vspace{-1.5cm}
\centering
\includegraphics[angle=0,origin=c,width=\textwidth]{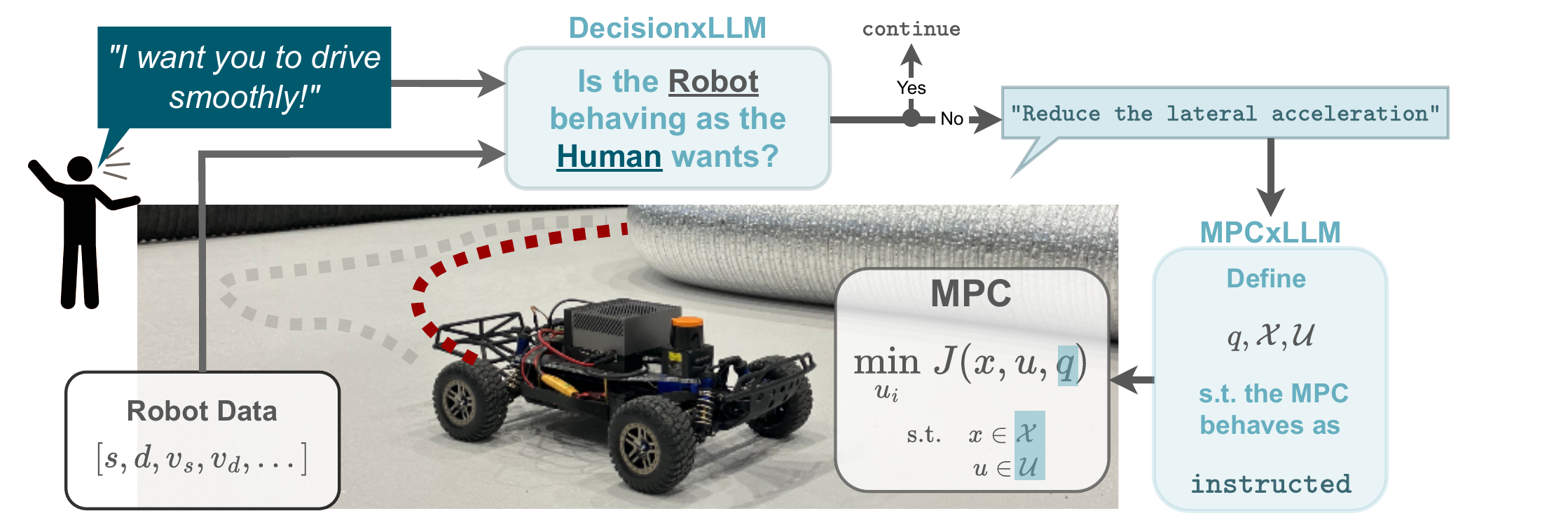}
\vspace{-0.5cm}
\captionof{figure}{Schematic overview of the proposed \gls{llm}-enhanced \gls{ads}. The \gls{llm} enables natural language-based \gls{hmi}, with a \textit{DecisionxLLM} stage analyzing robotic state information to ensure alignment with human preferences. If misalignment is detected, \textit{DecisionxLLM} instructs the \textit{MPCxLLM} stage to adjust the cost $J(x,u,q)$ and constraint ($\mathcal{X, U}$) parameters $x,u,q$ of a low-level \gls{mpc} controller, where safety and constraint satisfaction is ensured through the \gls{mpc}, while task adaption and decision-making are managed by a \gls{llm}.}
\label{fig:graphical_abstract}
\vspace{-0.5cm}
\end{strip}

\glsresetall

\begin{abstract}
\glspl{nn} trained through supervised learning, struggle with managing edge-case scenarios common in real-world driving due to the intractability of exhaustive datasets covering all edge-cases, making knowledge-driven approaches, akin to how humans intuitively detect unexpected driving behavior, a suitable complement to data-driven methods.
This work proposes a hybrid architecture combining low-level \gls{mpc} with locally deployed \glspl{llm} to enhance decision-making and \gls{hmi}. The \textit{DecisionxLLM} module evaluates robotic state information against natural language instructions to ensure adherence to desired driving behavior. The \textit{MPCxLLM} module then adjusts \gls{mpc} parameters based on \gls{llm}-generated insights, achieving control adaptability while preserving the safety and constraint guarantees of traditional \gls{mpc} systems.
Further, to enable efficient on-board deployment and to eliminate dependency on cloud connectivity, we shift processing to the on-board computing platform: We propose an approach that exploits \gls{rag}, \gls{lora} fine-tuning, and quantization. Experimental results demonstrate that these enhancements yield significant improvements in reasoning accuracy by up to 10.45\%, control adaptability by as much as 52.2\%, and up to 10.5$\times$ increase in computational efficiency (tokens/s), validating the proposed framework's practicality for real-time deployment even on down-scaled robotic platforms.
This work bridges high-level decision-making with low-level control adaptability, offering a synergistic framework for knowledge-driven and adaptive \gls{ads}.
\end{abstract}

\IEEEpeerreviewmaketitle

\section{Introduction} \label{sec:intro}

In the early 2010s~\cite{levinson2011towards,berger2012engineering}, it was widely anticipated by experts that research in \gls{ads} would soon lead to the widespread adoption of fully autonomous vehicles, fundamentally transforming the automotive sector. However, progress toward full autonomy proved more challenging than initially predicted.

Before the adoption of \gls{ml}, many autonomous driving systems were primarily addressed using classical robotic algorithms for perception, planning, and control, adhering to the \textit{See-Think-Act} cycle \cite{siegwart_amr}. While strongly principled, these approaches exhibited significant sensitivity to heuristics and parameter tuning. With the advent of \gls{ml} and especially \glspl{nn}, the ability to implicitly learn heuristics and improve robustness against parameter sensitivity was demonstrated \cite{nn_heuristics0, nn_heuristics1}. Consequently, efforts were directed towards substituting individual components of the \textit{See-Think-Act} cycle with \glspl{nn} or bypassing the cycle entirely through end-to-end learning paradigms, such as \gls{rl} \cite{rl_works0, rl_works1}.

Nowadays data-driven \gls{ml} approaches remain the predominant methodology in \gls{ads} ~\cite{geiger2013vision,nuscenes}. However, despite the considerable progress achieved through these approaches, full autonomy remains elusive. \gls{ml} systems inherently rely on extensive amounts of training data to generalize effectively, but edge-case scenarios are typically underrepresented in datasets. Consequently, data-driven approaches struggle in these contexts, requiring human intervention to address situations where little or no relevant data exists \cite{pavone2024gtc, heidecker2021application, bolte2019towards}.

These limitations suggest that driving is not solely a data-driven problem but partially relies on knowledge-driven reasoning \cite{wen2023dilu}. For example, when encountering anomalous scenarios, data-driven solutions must have been explicitly trained on such examples \cite{heidecker2021application}, whereas human drivers rely on common sense and situational reasoning to handle these situations effectively \cite{wen2023dilu}. Furthermore, the simulation or synthetic generation of every possible peculiar road scenario is intractable, highlighting the necessity for knowledge-driven methodologies in \gls{ads}.

In recent years, significant advancements have been made in \glspl{llm}, which represent the closest approximation to artificial knowledge systems to date~\cite{achiam2023gpt,bai2023qwen,abdin2024phi}. While \glspl{llm} have demonstrated their capabilities in robotic tasks such as manipulation and scene understanding \cite{duan2024manipulate, wang2024embodiedscan}, their adoption in the \gls{ads} domain remains relatively limited especially because existing robotic embodied \gls{ai} systems predominantly depend on cloud-based models, such as \emph{GPT4} \cite{achiam2023gpt}. However, reliance on cloud infrastructure introduces concerns regarding latency, connection stability, security, and privacy \cite{iot_good0, iot_good1, slm}. As a result, local deployment of \glspl{llm} on robotic platforms emerges as a more robust and secure alternative for \gls{ads}.

Concerns persist regarding the deployment of \glspl{llm} for critical tasks, given their susceptibility to hallucinations \cite{huang2023surveyhallucination}. Within this work, their integration into every facet of autonomous driving is neither suggested nor advisable. Instead, emphasis is placed on leveraging their knowledge-driven reasoning capabilities in specific scenarios where their strengths are most applicable.

Hence, a hybrid system architecture is proposed that adheres to the classical \emph{See-Think-Act} cycle while incorporating a locally deployed, knowledge-based \gls{llm}. The architecture in this work is designed to enable \glspl{llm} to support \gls{hmi}, decision-making, and dynamic control adjustments, while the evaluation has been performed on a 1:10 scaled autonomous car platform \cite{forzaeth}. The underlying controller operates based on a low-level \gls{mpc}, ensuring safety through constraint satisfaction. A \textit{DecisionxLLM} module monitors robotic state data sampled over recent time intervals, analyzing adherence to user instructions. If discrepancies are detected, the \textit{MPCxLLM} stage interacts with the \gls{mpc} controller, adjusting cost function weights and constraints as needed. This approach facilitates seamless \gls{hmi} while maintaining the safety and reliability inherent in \gls{mpc}-based systems. The proposed framework allows switching between \glspl{llm}; for instance, \textit{GPT4o} could be utilized for tasks requiring extensive cloud-based resources, while local \glspl{llm} such as \textit{Qwen2.5-7b} can be deployed depending on connectivity constraints and other operational requirements.

To summarize, the contributions of this work are as follows:
\begin{enumerate}[I] 
    \item \textbf{Knowledge-based Decision Making:} A framework is proposed for leveraging \glspl{llm} to interpret robotic data conditioned on human desired driving behavior, enabling decision-making based on behavioral adherence. By implementing the proposed \gls{rag} and \gls{lora} fine-tuning techniques, decision-making accuracy is improved by up to 10.45\% on local \glspl{llm}. Open-source code: \href{https://github.com/ForzaETH/LLMxRobot}{github.com/ForzaETH/LLMxRobot}.
    
    \item \textbf{Human-Machine Interaction:} Adherence is identified in relation to human prompts, enabling natural language-based \gls{hmi} through dynamic adjustments of cost and constraint parameters in the low-level \gls{mpc} controller. This approach enables an increase in control adaptability by up to 52.2\%.
    
    \item \textbf{Embodied AI on the Edge:} The proposed framework avoids reliance on cloud services by deploying \glspl{llm} locally on embedded platforms, such as the \textit{Jetson Orin AGX}, ensuring reliability, enhanced privacy, and improved security on computationally constrained devices. By employing \texttt{Q5\_k\_m} quantization and the \texttt{llama.cpp} inference engine, up to a 10.5-fold increase in computational efficiency (tokens/s) can be achieved on embedded \glspl{obc}.
\end{enumerate}

\section{Related Work} \label{sec:relatedwork}

This section reviews relevant work on the use of \glspl{llm} for robotic control (\Cref{subsec:rw_llm_control}) and decision-making (\Cref{subsec:rw_llm_decision}), concluding with a contextual summary (\Cref{subsec:rw_summary}).

\subsection{LLMs and Robot Control} \label{subsec:rw_llm_control}

Recent studies highlight that direct control of robotic actuators by \glspl{llm} is unsuitable due to their lack of training data on actuator-level commands, incompatibility with real-time control frequencies, and limited suitability for classical control paradigms \cite{l2r}. Instead, approaches such as \cite{l2r, ismail2024narrate, ma2023eureka} emphasize the role of reward functions as intermediaries, enabling interaction between \glspl{llm} and low-level controllers like \gls{mpc}. Here, the \gls{llm} interacts with the cost function and system constraints, allowing for interpretable and flexible control adaptation.

Building on this, \citet{ismail2024narrate} propose using an \gls{llm} to generate objective functions and constraints for manipulation tasks based on human prompts. This architecture combines the adaptability of \glspl{llm} with the safety and constraint guarantees of classical \gls{mpc} controllers.

Similarly, \citet{ma2023eureka} demonstrates the significance of reward functions within \gls{llm}-\gls{rl} interactions. Instead of focusing on cost functions and constraints as in \cite{l2r, ismail2024narrate}, the \gls{llm} iteratively designs reward functions and domain randomization strategies during the training of an \gls{rl} locomotion policy, this has shown to yield great flexibility in designing effective \textit{Sim-to-Real} policies \cite{ma2023eureka}. 

Collectively, these works demonstrate that \glspl{llm} are more effective at interpreting and adjusting reward functions and constraints rather than acting as low-level controllers. This approach combines the interpretability and flexibility of \glspl{llm} with the safety guarantees of traditional controllers. However, existing solutions depend on cloud-based \emph{GPT4} models, introducing concerns regarding privacy, latency, and internet reliability \cite{iot_good1, iot_good0, slm}. Additionally, these approaches limit the \glspl{llm} role to adapting the behavior of the low-level controller rather than actively participating in any decision-making processes.

\subsection{Robotic Decision Making with LLMs} \label{subsec:rw_llm_decision}
\citet{wen2023dilu} propose the \textit{DiLU} framework, which utilizes an \gls{llm} for decision-making in autonomous driving. \textit{DiLU} incorporates three core components: a reasoning module that interprets the current driving scenario and generates high-level decisions, a reflection module that evaluates and refines these decisions based on previous outcomes, and a memory module that accumulates experiences from previous interactions. This architecture enables \textit{DiLU} to integrate reasoning with iterative refinement, allowing it to handle driving scenarios effectively.

While \textit{DiLU} demonstrates reasoning and adaptability capabilities, it operates within a discrete action space, which limits its applicability to continuous control tasks common in real-world robotics. Furthermore, it relies on cloud-based \emph{GPT4} models, with the aforementioned cloud-reliance downsides \cite{iot_good0, iot_good1, slm}. Lastly, the framework has only been validated in simulation within a simple highway lane-switching \gls{rl} environment. It remains to be seen whether this approach can generalize effectively to physical robotic systems operating in real-world conditions.

\subsection{Summary of LLMs and Robotics} \label{subsec:rw_summary}
As summarized in \Cref{tab:rw_summary}, existing works like \citet{l2r} and \citet{ismail2024narrate} focus on manipulation and locomotion tasks, relying on classical low-level controllers and lacking robotic reasoning. \textit{DiLU} explores \gls{llm}-based decision-making for autonomous driving but is limited to a discrete action space in a simulated highway lane-switching environment, restricting its applicability from real-world continuous control tasks. Additionally, these approaches depend on cloud-based \emph{GPT4}, posing challenges with latency, privacy, and connectivity \cite{iot_good0, iot_good1, slm}.

In contrast, our proposed approach emphasizes local edge deployment of the \gls{llm}, and decision-making in a continuous action space on a physical robotic car, directly grounded on robotic sensor data.

\begin{table}[!htb] 
    \centering 
    \resizebox{\columnwidth}{!}{%
    \begin{tabular}{l|c|c|c|c}
    \toprule
    \textbf{Related Work} & \textbf{Task} & \textbf{Decision} &  \textbf{Controls} & \textbf{\gls{llm}} \\
    \midrule
    \citet{ismail2024narrate} & Manipulation & \xmark & \cmark & GPT4$\dagger$ \\
    \citet{l2r} & Locomotion & \xmark & \cmark & GPT4$\dagger$ \\
    \citet{ma2023eureka} & Locomotion & \xmark & \cmark & GPT4$\dagger$ \\
    \citet{wen2023dilu} & Driving & \cmark & \xmark & GPT4$\dagger$ \\
    \textbf{ours} & Driving & \cmark & \cmark & Qwen-7b  \\
    \bottomrule
    \end{tabular}%
    }
    \caption{Comparison of related work integrating \glspl{llm} with robotics, highlighting differences in tasks, decision-making, control influence, and \gls{llm} models used. $\dagger$ denotes cloud dependency.}
    \label{tab:rw_summary}
\end{table}

\section{Methodology} \label{sec:methodology}
This section describes the hardware and robotic platform used for onboard \gls{llm} processing in \Cref{subsec:hw}. \Cref{subsec:decisionxllm} describes how the decision-making module \textit{DecisionxLLM} utilizes robotic data for reasoning and integrates with the overall stack. Finally, \Cref{subsec:mpcxllm} explains the interaction between the \gls{llm} and the controller within \textit{MPCxLLM}, including the \gls{mpc} formulation.

\subsection{Robotic System and Computational Hardware} \label{subsec:hw}
\begin{figure}[!htb]
    \centering
    \includegraphics[width=\columnwidth, trim={0 2cm 0 1cm}, clip]{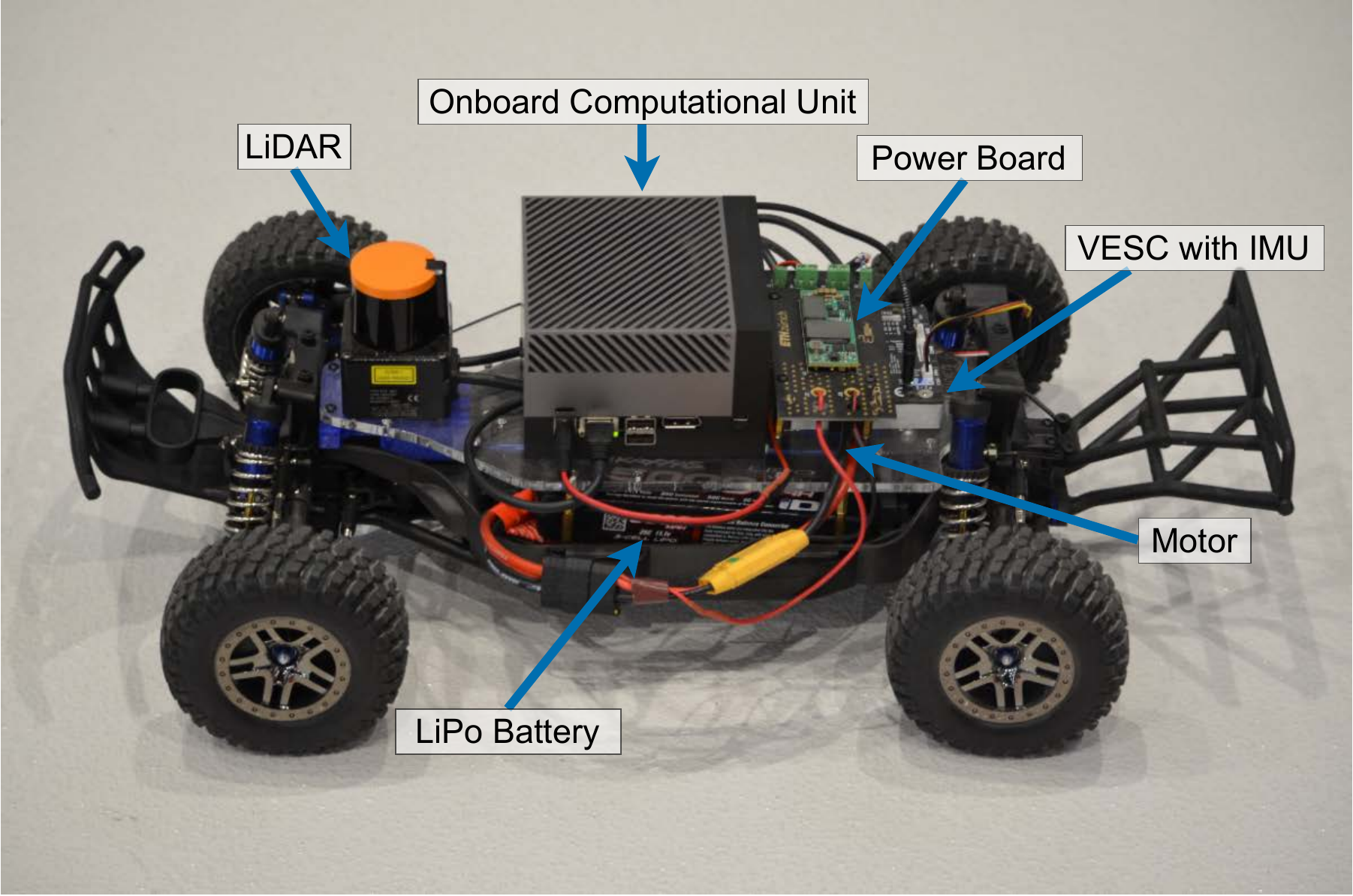}
    \caption{The 1:10 scaled robotic platform, utilizing the \textit{Jetson Orin AGX} as the \gls{obc} for executing computations related to the locally deployed \gls{llm} and the \gls{ads} autonomy stack.}
    \label{fig:hardware}
\end{figure}

In this work, the robotic platform of \Cref{fig:hardware}, along with its autonomy algorithms as detailed in the open-source \textit{F1TENTH} autonomy stack \cite{forzaeth}, is employed. The autonomy stack has been extended with the integration of a kinematic \gls{mpc} controller, further described in \Cref{subsec:mpc}. Consequently, positional state information is represented in the \textit{Frenet} coordinate frame, where the $s$ coordinate indicates longitudinal progress along a global trajectory, and the $d$ coordinate denotes lateral deviation from that trajectory, following the conventions in \cite{forzaeth}.

A key hardware component is the \textit{Jetson Orin AGX} serving as the \gls{obc}. This \gls{obc} incorporates a 2048-core NVIDIA Ampere architecture \gls{gpu} with 64 Tensor Cores, delivering 275 TOPS, and is utilized for \gls{llm} inference. Additionally, the \gls{cpu}, a 12-core Arm Cortex-A78AE, is responsible for running the autonomy stack, including the \gls{mpc}. The \gls{obc} is equipped with 64GB of shared \gls{ram}, providing ample memory for computational tasks.

In this work, \textit{GPT4o} is utilized via the \textit{OpenAI} API \cite{hurst2024gpt4o} as a cloud-based upper baseline, while two locally deployable models, Phi-3-mini-4k-instruct\footnote{https://huggingface.co/unsloth/Phi-3-mini-4k-instruct} (referred to as Phi3-mini), a 3B parameter \gls{llm}, and Qwen2.5-7b-Instruct\footnote{https://huggingface.co/unsloth/Qwen2.5-7B-Instruct}, a 7B parameter \gls{llm}, are sourced directly from \textit{HuggingFace}. These models were intentionally selected to validate the robustness and adaptability of the proposed framework, demonstrating its effectiveness across diverse architectures and parameter scales.

\subsection{DecisionxLLM --- Decision Making with Robotic Data} \label{subsec:decisionxllm}
\begin{figure}[!htb]
    \centering
    \includegraphics[width=\columnwidth]{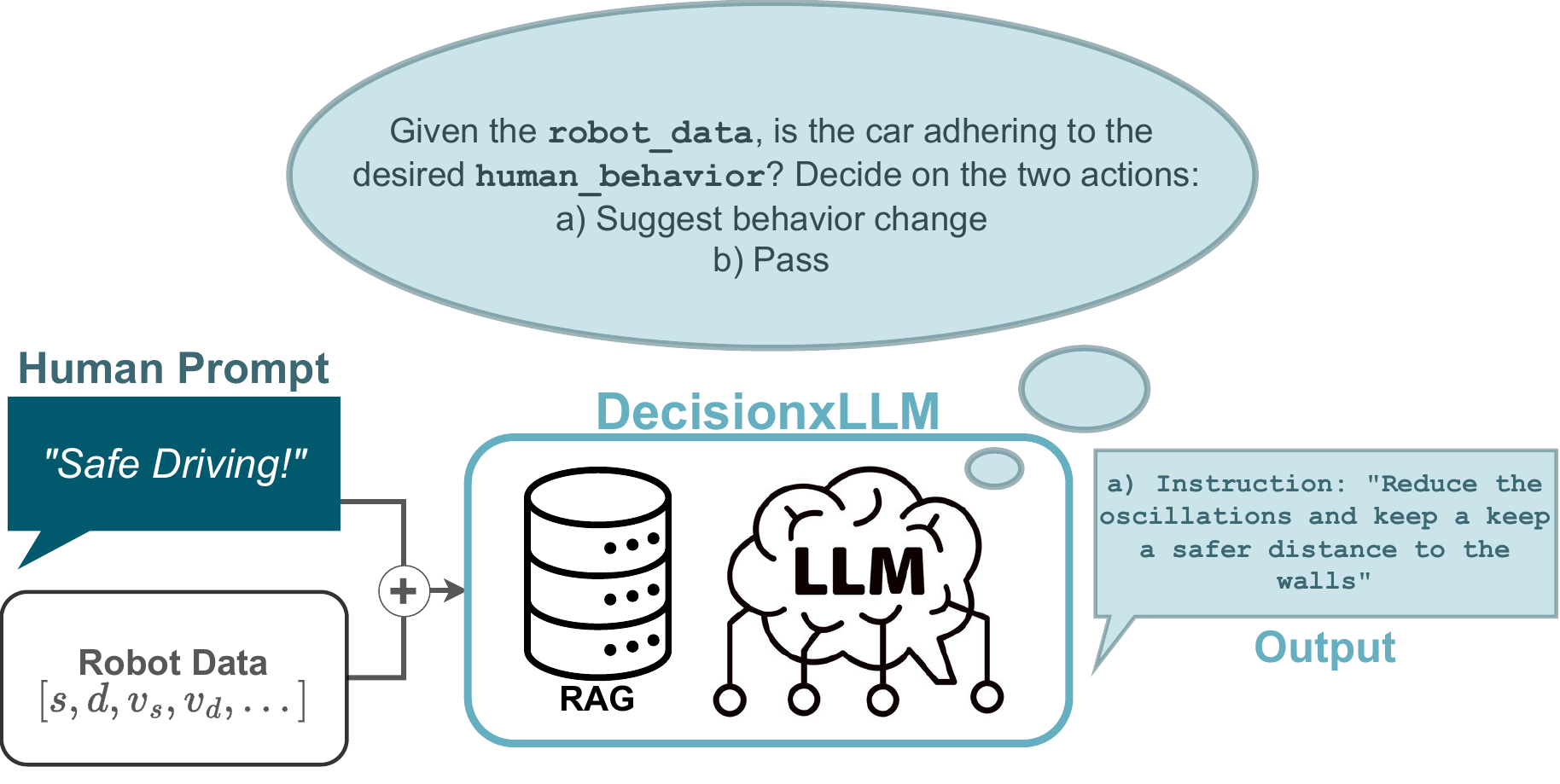}
    \caption{Diagram illustrating the decision-making process of the \gls{llm}, where it evaluates robot data conditioned on a desired driving behavior based on a human prompt. The \gls{llm} determines whether the behavior aligns with expectations or suggests necessary adjustments.}
    \label{fig:decision_llm}
\end{figure}

The decision-making mechanism of \textit{DecisionxLLM}, shown in \Cref{fig:decision_llm}, enables dynamic evaluation of robotic data against desired driving behavior expressed through natural language prompts. The system processes a brief temporal snapshot (e.g., 2 seconds) of the robot's state, including position, velocity, and proximity to environmental boundaries.

Given this robotic data and a human-defined driving behavior prompt, the \gls{llm} assesses whether the robot adheres to the specified behavior. For example, if a passenger in an autonomous taxi requests a smoother ride due to discomfort, the \gls{llm} could infer a reduction in lateral acceleration and prioritize gentle maneuvers, enhancing overall user experience through natural language \gls{hmi}.

The \gls{rag} module within the \textit{DecisionxLLM} architecture, inspired by \cite{wen2023dilu}, optionally enhances the system by enabling memory modules to enrich the prompt with relevant context. This includes safety-critical and robot-specific information, such as nominal operating ranges (e.g., speed limits, distance thresholds). This capability allows human users to define custom safety and preference profiles while significantly improving the \gls{llm}'s decision-making abilities by augmenting robot-specific constraints into the prompt. This augmentation is particularly valuable on computationally constrained embedded \glspl{obc}, as the performance improvement comes without having to employ larger compute heavy \glspl{llm}. An example of the decision \gls{rag} is provided in \Cref{appendix:decision}, \Cref{lst:decision_rag}.

If the \textit{DecisionxLLM} determines that the robot behavior aligns with the desired behavior, no further action is taken. However, if deviations are detected, the module generates a concise adjustment instruction in natural language, specifying how the behavior should be corrected. This instruction seamlessly integrates with the \textit{MPCxLLM} module, where it serves as input to dynamically adjust relevant parameters, ensuring alignment with the desired behavior.

\subsection{MPCxLLM --- Controller Interaction} \label{subsec:mpcxllm}
\begin{figure}[!htb]
    \centering
    \includegraphics[width=\columnwidth]{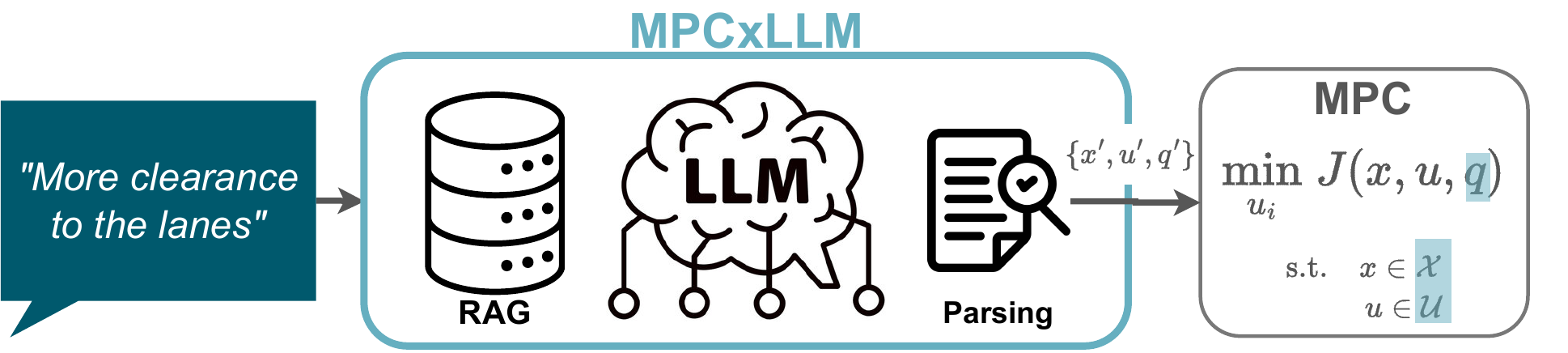}
    \caption{Illustration of the \textit{MPCxLLM} architecture: a natural language instruction serves as input, optionally enhanced by a \gls{rag}, processed by the \gls{llm}, and finally parsed to extract relevant parameters, which are then transmitted to the \gls{mpc} via \gls{ros} dynamic reconfigure.}
    \label{fig:mpc_llm}
\end{figure}

The interaction between the \textit{MPCxLLM} controller and the \gls{llm}, depicted in \Cref{fig:mpc_llm}, follows principles from \cite{l2r, ismail2024narrate}. This integration enables an \gls{llm}, aware of the \gls{mpc} formulation and its adjustable parameters, to interface with the low-level controller. As a result, task flexibility is achieved through natural language-based \gls{hmi}, while the \gls{mpc} ensures safety and constraint satisfaction at the low level.

Importantly, the inference latency of the \gls{llm} is decoupled from the control frequency of the \gls{mpc}. Operating at a higher abstraction level, the \gls{llm} intermittently adjusts the \gls{mpc} parameters without interfering with the \gls{mpc} fixed-frequency control loop. This ensures control stability and safety, while the \gls{llm} focuses on task-level adaptations.

Similar to the \gls{rag} module described in \Cref{subsec:decisionxllm}, an optional \gls{rag} can be integrated here. Given the computational constraints of the \gls{obc} and the limited size of the locally deployed \gls{llm}, the \gls{rag} can significantly improve performance by enriching prompts with context-specific hints tailored for the \gls{mpc}. An example of such a \gls{mpc} \gls{rag} memory is provided in \Cref{appendix:mpc}, \Cref{lst:mpc_rag}. It is worth noting that the \gls{rag} module is optional and can be disabled, as evaluated in \Cref{tab:control_res}.

The \gls{mpc} is based on a kinematic model, with its cost function primarily designed to ensure accurate tracking of the given trajectory and velocity.

\subsubsection{Kinematic Model} \label{subsec:mpc}
\begin{equation}
\begin{aligned}
\dot{s} &= \frac{v \cos(\Delta \phi)}{1 - \kappa_{r}(s) \, n}, \\
\dot{n} &= v \sin(\Delta \phi), \\
\Delta \dot{\phi} &= \frac{v \tan(\delta)}{L}
  - \kappa_{r}(s) \cdot \frac{v \cos(\Delta \phi)}{1 - \kappa_{r}(s) \, n}.
\end{aligned}
\end{equation}
where $\kappa_{r}$ indicates the curvature of the reference trajectory. $L$ is the distance from the front axle to the rear axle. $s$ denotes the distance traveled along the reference trajectory, and $n$ is the lateral deviation from this trajectory. $\Delta \phi$ represents the heading angle error. $\delta$ denotes the steering angle and $v$ is the velocity. The incremental model is utilized to smooth the control inputs. Consequently, the state of the model is comprised of the following five variables:
\begin{align}x=\begin{bmatrix} s &n &\Delta \phi & \delta & v \end{bmatrix}^{T}\end{align}

The input variables include the steering angle difference $\Delta \delta$ and the longitudinal acceleration $a$:
\begin{align}u=\begin{bmatrix} \Delta \delta & a \end{bmatrix}^{T}\end{align}

\subsubsection{MPC Formulation} \label{subsec:mpc_form}
\begin{align}
\mathop \text{min}\limits_{u_{k+i|k}}& J(x, u, q) = \sum^{N-1}_{i=0} q_n \cdot n_{k+i|k}^2+q_v \cdot (v_{k+i|k}-v_{ref})^2 \notag\\&+q_{\alpha} \cdot \Delta \phi_{k+i|k}^2 +\Vert \Delta u_{k+i|k} \Vert_{q_R}+q_n \cdot n_{k+N|k}^2 \notag 
 \\& + q_v \cdot (v_{k+N|k}-v_{ref})^2 +q_{\alpha} \cdot \Delta \phi_{k+N|k}^2 \\
s.t. \notag \\
& {x}_{k+1+i|k}=f(x_{k+i|k}, u_{k+i|k}) \notag\\
&x_{k+i|k}\in \mathcal{X}, u_{k+i|k}\in \mathcal{U}\\
&\forall i=0,1,\cdots,N-1 \notag
\end{align}
where $N$ is the prediction horizon. $q_n$, $q_{\alpha}$, $q_v$ are state weight parameters. $q_R$ represents the weight matrix for the control inputs. Its diagonal elements respectively indicate the penalties on the difference of the steering angle and the longitudinal acceleration, denoted by $q_{\Delta \delta}$ and $q_{ac}$. $\mathcal{X}$ and $\mathcal{U}$ represent the sets of constraints for the states and inputs, respectively. Among them, the velocity $v$ and the steering angle $\delta$ are each limited in magnitude. The vehicle's lateral error is constrained within the road boundaries, and an online tuning parameter \(\epsilon\) is introduced as a boundary inflation factor to ensure the vehicle’s driving safety. The steering angle difference and the longitudinal acceleration are also limited in their respective ranges. To simplify the expression of the optimization problem, the lateral acceleration constraint $a_c(x,u)$ based on the vehicle's kinematic model is not explicitly included in the formulation. 

Overall, the weight parameters $q$, the constraints on states 
$\mathcal{X}$ and inputs $\mathcal{U}$, and the boundary inflation $\epsilon$, can all be treated as adjustable parameters for the \gls{llm} to tune so that the vehicle can exhibit the desired behavior.

\subsection{LoRA Finetuning of the LLMs} \label{subsec:finetuning}
While \textit{GPT4o} together with robot or task-specific RAG have shown to be zero-shot capable, small \glspl{llm} which would be locally deployable perform significantly worse than the large cloud-bound \textit{GPT4o} and thus have to be \gls{lora} fine-tuned with synthetic data derived from \textit{GPT4o}. Although the correctness of the synthetic data is not formally guaranteed, observed response quality was deemed high enough for \gls{lora} fine-tuning on smaller \glspl{llm}. Hence, finetuning via \gls{peft} methods, particularly \gls{lora} \cite{lora}, is employed. \gls{lora} can reduce the number of trainable parameters by thousands of times and the \gls{gpu} memory requirement by up to threefold, significantly simplifying the finetuning process \cite{lora}. For \gls{lora}-based \gls{peft}, the \texttt{unsloth} framework is utilized \cite{unsloth}. Training utilizes synthetic datasets generated by zero-shot prompting \emph{GPT-4o} \cite{hurst2024gpt4o} across each domain:

\begin{enumerate}[I] 
    \item \textbf{Synthetic Data for DecisionxLLM:} \emph{GPT4o} generates state summaries using randomized parameters derived from the decision-\gls{rag}, detailed in \Cref{lst:decision_rag}. By leveraging randomized nominal operation ranges and thresholds (e.g., speed ranges, critical distances to boundaries, etc.), \emph{GPT4o} synthesizes robot state representations and identifies deviations from expected behavior according to the \gls{rag}. These randomized parameters encourage the local \gls{llm} to focus more emphasis on the decision-\gls{rag} hints. This enhances the robot's ability to ground its decision-making process in the provided hints, facilitating customization of safety and preference profiles by the user. More information in \Cref{appendix:lora}.

     \item \textbf{Synthetic Data for MPCxLLM:} \emph{GPT4o} generates data using randomized parameters in a base \gls{mpc} formulation (\Cref{lst:base_mpc}). The model adapts these parameters to enforce specific driving behaviors while providing brief justifications. This dataset enables \gls{lora} to learn interactions between \gls{mpc} elements, with training data designed to be out-of-distribution due to the parameter randomization and an altered \gls{mpc} formulation being employed during inference. More information in \Cref{appendix:lora}.
\end{enumerate}

Post-training quantization is applied to enhance inference speed and efficiency by converting the finetuned \glspl{llm} into the \texttt{Q5\_k\_m GGUF} format. This compression reduces both memory usage and computational demands, enabling notably faster inference when utilizing the \texttt{llama.cpp} inference engine \cite{chavan2024faster, llamacpp} on the resource-constrained \gls{obc}, as demonstrated in \Cref{tab:compute}.

This combined approach ensures efficient training, optimized inference, and enhanced onboard decision-making capabilities for both \textit{MPCxLLM} and \textit{DecisionxLLM} modules.

\section{Results} \label{sec:results}
This section presents the experimental setup, along with qualitative and quantitative results of the \textit{DecisionxLLM} modules's ability to assess whether the vehicle adheres to instructed driving behaviors based on robotic state information in \Cref{subsec:results_decisionxllm}. Next, the controller adaptability through \textit{MPCxLLM} is evaluated in \Cref{subsec:results_mpcxllm}. Finally, the computational performance of the edge-deployed \gls{llm} is evaluated on the resource-constrained \gls{obc} in \Cref{subsec:results_compute}.

\subsection{Reasoning on Robotic Observations} \label{subsec:results_decisionxllm}

\Cref{bubble:1a} illustrates a qualitative example of the \textit{DecisionxLLM} output. The \texttt{human\_prompt} instructs the system to \emph{"Drive normally"}. The \gls{llm} evaluates the robot state against this directive, leveraging optional \texttt{rag\_hints} that provide contextual information, such as for example nominal vehicle speed and safe distances from track boundaries. Detailed \texttt{rag\_hints} are provided in \Cref{lst:decision_rag}.

In this scenario, the \texttt{rag\_hints} define "normal" driving as minimal lateral oscillations, safe clearance from boundaries, and sufficient speed. The \gls{llm} identifies anomalies within the robot state: significant lateral oscillations, proximity to the right wall at \SI{0.1}{\metre}, and a low speed of \SI{1}{\metre\per\second}. Based on these observations, the module concludes that the state does not conform to the human directive of \emph{"driving normally"}.

Subsequently, the \gls{llm} generates a natural language instruction for the \textit{MPCxLLM} module to correct these issues, enhancing interpretability and ensuring alignment with the desired driving behavior.

\begin{decisionbubble}{1a}{Human Instruction}
Drive normally!
\end{decisionbubble}

\begin{decisionbubble}{1b}{Prompt}
The human wants to: \texttt{human\_prompt}.\\
The data has been sampled for 2.0 seconds in 4 samples.   
\begin{itemize}
    \item s-coordinate: \texttt{19, 20, 21, 22}
    \item d-coordinate: \texttt{-0.6, 0.6, -0.65, 0.61}
    \item s-speed: \texttt{1.0, 1.0, 0.9, 1.1}
    \item d-speed: \texttt{1.2, -1.2, 1.21, -1.2}
    \item distance to left wall: \texttt{1, 1, 1, 1}
    \item distance to right wall: \texttt{0.1, 0.1, 0.1, 0.1}
    \item crashed: \texttt{False}
\end{itemize}
Here are some guides to help you reason: \texttt{rag\_hints}.\\
Check if the car is doing what the human wants. Choose one of the following actions to command the car: 
\begin{itemize}
    \item a) Continue behavior
    \item b) Change behavior: \texttt{instruction}
\end{itemize}
\end{decisionbubble}

\begin{decisionbubble}{1c}{Response}
The car is driving with s-speeds close to 1.0 m/s, which is \textbf{below the normal driving speed} range of 5-7 m/s. The d-coordinate oscillates between positive and negative values, indicating \textbf{oscillation around the racing line}. \textbf{The d-speed is high}, exceeding 0.5 m/s, which is not normal. The car is not crashed, facing the wall, or reversing. The distances to the \textbf{walls are very close}, which is unsafe. The car is \textbf{not adhering to the human's} desired driving style of "drive normally."

\texttt{Action:} \textbf{b) Change behavior}

\texttt{Instruction:} The car should \textbf{increase its s-speed} to a normal range of 5-7 m/s, \textbf{reduce the oscillation} in d-coordinate, and \textbf{move closer to the centerline to increase safety}.
\end{decisionbubble}
\setcounter{bubblegroup}{1}

On a quantitative note, the \textit{DecisionxLLM} module is evaluated based on its ability to determine whether a robotic state adheres to the human's desired driving behavior. A dataset containing 200 sets of robot states was curated and labeled according to 8 example human prompts, allowing for quantitative evaluation. Example prompts include:
\begin{enumerate}[I]
    \item "Drive faster than 3 m/s!"
    \item "Normal driving on the racing line."
    \item "Reverse the car!"
\end{enumerate}

These prompts were selected for their suitability to programmatically classify the robot states based on the predefined driving characteristics listed above. The \gls{llm} performs a binary classification over 1600 robot state samples, to determine adherence conditioned on a desired driving behavior. More information on the definition of decision-making accuracy in \Cref{appendix:decision}.

\begin{table}[!htb] 
    \centering 
    \begin{adjustbox}{max width=\columnwidth}
    \begin{tabular}{l|c|c|c|c|c}
    \toprule
    \textbf{LLM} & \textbf{Params} & \textbf{Quant} & \textbf{LoRA} & \textbf{RAG} & \textbf{Accuracy [\%]}\bm{$\uparrow$}\\
    \midrule
    GPT4o & ? &  ? & \xmark & \xmark & \textbf{81.68} \\
    Phi3-mini & 3.8B & FP16 & \xmark & \xmark & 72.15 \\
    Qwen2.5 & 7B & FP16 & \xmark & \xmark & 77.75\\
    \midrule
    GPT4o & ? & ? & \xmark & \cmark & \textbf{92.48} \\
    Phi3-mini & 3.8B & FP16 & \xmark & \cmark & 78.69 \\
    Qwen2.5 & 7B & FP16 & \xmark & \cmark & 82.47 \\
    \midrule
    Phi3-mini & 3.8B & FP16 & \cmark & \cmark & 82.60 \\
    Qwen2.5 & 7B & FP16 & \cmark & \cmark & \textbf{87.32} \\
    \midrule
    Phi3-mini & 3.8B & Q5 & \cmark & \cmark & 84.95 \\
    Qwen2.5 & 7B & Q5 & \cmark & \cmark & \textbf{87.02} \\
    \bottomrule
    \end{tabular}%
    \end{adjustbox}
    \caption{Decision-making accuracy across various \glspl{llm}, illustrating the impact of quantization, \gls{lora} fine-tuning, and \gls{rag} on the performance of the \textit{DecisionxLLM} module in evaluating whether a robotic state aligns with desired driving behavior. Higher accuracy values indicate improved adherence to the specified driving characteristics. ? indicates that this is proprietary knowledge, not known to the public.}
    \label{tab:decision_res}
\end{table}

\Cref{tab:decision_res} demonstrates the performance of the \textit{DecisionxLLM} module in determining whether a robotic state adheres to the desired human driving behavior. All performance improvements are stated in absolute percentage points. The results indicate that the inclusion of \gls{rag} consistently enhances model performance across all tested \glspl{llm}, with an average improvement of 7.35\% (\textit{GPT4o/Phi3/Qwen2.5} -- +10.8/+6.54/+4.72\%). 
Fine-tuning locally deployable \glspl{llm} via \textit{GPT4o} distillation, as detailed in \Cref{subsec:finetuning}, further improves model accuracy by an average of 4.38\% (\textit{Phi3/Qwen2.5} -- +3.91/+4.85\%). 
Lastly, quantization, which is essential for deployment on computationally constrained \glspl{obc}, does not substantially degrade decision-making performance, maintaining accuracy within a margin of 1.02\% relative to full-precision models (\textit{Phi3/Qwen2.5} -- +2.35/-0.3\%).

In summary, the proposed system effectively reasons over robotic state information conditioned on human driving instructions, benefiting from \gls{rag}, \gls{lora} fine-tuning (\textit{Phi3/Qwen2.5} -- +10.45/+9.57\%), and demonstrating resilience to the effects of model quantization for local deployment. Further, comparing the two local \glspl{llm}, \textit{Qwen2.5-7b} consistently outperforms \textit{Phi3-mini-3.8b} across all settings, making it the preferred choice. That said, \textit{GPT4o} achieves the highest performance overall, and the framework's flexibility enables its use in scenarios where dependency on cloud connectivity is warranted.

While \Cref{tab:decision_res} presents the average decision-making accuracy on the custom dataset, \Cref{tab:ext_decision_res} provides a detailed breakdown across sub-categories and additional highly contemporary \glspl{llm}, including a \emph{DeepSeek R1} distilled \emph{Qwen2.5-7b} model \cite{r1}.

\begin{table*}[!htb] 
    \centering 
    \begin{tabular}{l|c|c|c|c|c|c|c|c|c}
    \toprule
    \textbf{LLM} & \textbf{Params} & \textbf{Quant} & \textbf{LoRA} & \textbf{RAG} & \bm{$E_C [m]\downarrow$} & \bm{$E_V [ms^{-1}]\downarrow$} &  \bm{$E_R [ms^{-1}]\downarrow$} & \bm{$E_S [ms^{-2}]\downarrow$} & \textbf{Improve [\%]}\bm{$\uparrow$} \\
    \midrule
    \gls{mpc} (baseline) & - & - & - & - & 0.7 & 1.6 & 4.6 & 1.5 & - \\
    \midrule
    GPT4o & ? & ? & \xmark & \xmark & \textbf{0.5 (19.3\%)} & \textbf{1.8 (-11.8\%)} & 1.5 (68.1\%) & \textbf{1.4 (5.1\%)} & \textbf{20.2\%} \\
    Phi3-mini & 3.8B & FP16 & \xmark & \xmark & N.C. & 1.9 (-17.1\%) & 2.8 (39.5\%) & 11.8 (-711.4\%) & -229.6\%$^\dagger$ \\
    Qwen2.5 & 7B & FP16 & \xmark & \xmark & 0.7 (-14.6\%) & 1.8 (-14.4\%) & \textbf{1.3 (70.9\%)} & 2.0 (-40.4\%) & 0.5\% \\
    \midrule
    GPT4o & ? & ? & \xmark & \cmark & \textbf{0.5 (21.3\%)} & 0.3 (81.5\%) & \textbf{1.2 (74.7\%)} & 1.3 (7.5\%) & \textbf{46.3\%} \\
    Phi3-mini & 3.8B & FP16 & \xmark & \cmark & 0.6 (6.6\%) & 0.6 (60.8\%) & 4.2 (9.8\%) & \textbf{0.3 (82.5\%)} & 39.9\% \\
    Qwen2.5 & 7B & FP16 & \xmark & \cmark & 0.6 (8.6\%) & \textbf{0.1 (91.7\%)} & 1.5 (67.7\%) & 1.4 (4.8\%) & 43.2\% \\
    \midrule
    Phi3-mini & 3.8B & FP16 & \cmark & \cmark & 0.7 (-0.3\%) & 0.6 (59.3\%) & \textbf{1.0 (78.8\%)} & \textbf{0.8 (42.8\%)} & 45.1\% \\
    Qwen2.5 & 7B & FP16 & \cmark & \cmark & \textbf{0.4 (37.2\%)} & \textbf{0.4 (71.9\%)} & 1.1 (76.4\%) & 1.1 (23.2\%) & \textbf{52.2\%} \\
    \midrule
    Phi3-mini & 3.8B & Q5 & \cmark & \cmark & 0.57 (12.6\%) & 0.5 (70.7\%) & 1.2 (73.6\%) & 1.1 (-22.3\%) & 44.\% \\
    Qwen2.5 & 7B & Q5 & \cmark & \cmark & \textbf{0.43 (33.4\%)} & \textbf{0.2 (85.8\%)} & \textbf{1.3 (71.3\%)} & \textbf{1.5 (-3.3\%)} & \textbf{46.8\%} \\
    \bottomrule
    \end{tabular}
    \caption{Quantitative Comparison of \gls{llm} configurations with \gls{mpc}. Performance metrics include deviation from the centerline ($E_C [m]$), reference velocity ($E_V [ms^{-1}]$), reversing accuracy ($E_R [ms^{-1}]$), and driving smoothness ($E_S [ms^{-2}]$), with percentage improvements shown relative to the \gls{mpc} baseline. The average improvement column summarizes overall performance across all metrics. $\dagger$ indicates an average over completed runs, excluding N.C. (\textit{Not Completed}). ? indicates that this is proprietary knowledge, not known to the public.}
    \label{tab:control_res}
\end{table*}

\subsection{MPCxLLM --- Control Adaptability} \label{subsec:results_mpcxllm}
The \textit{MPCxLLM} module is evaluated, in a closed-loop simulation environment of the open-source \textit{F1TENTH} autonomy stack \cite{forzaeth}, where the \gls{mpc} handles vehicle control. In this context, simulation is preferred over physical testing due to time efficiency and access to ground-truth data.

The \gls{llm} governs interaction scenarios to enable the computation of quantifiable \gls{rmse} metrics. We assess changes in closed-loop behavior relative to the default \gls{mpc} formulation outlined in \Cref{subsec:mpc}, using the following criteria:

\begin{enumerate}[I] 
    \item \textbf{Centerline:} \gls{rmse} of the distance from the centerline, denoted as $E_C [m]$. Instruction prompt: "Drive as far away from the walls as possible!"
    \item \textbf{Reference Velocity:} \gls{rmse} deviation from a human-provided reference velocity, denoted as $E_V [ms^{-1}]$.  Instruction prompt: "Follow the reference velocity of 1.25 m/s as closely as possible!"
    \item \textbf{Reversing:} \gls{rmse} deviation from a negative reference velocity specified by the human, denoted as $E_R [ms^{-1}]$. Instruction prompt: "Drive the track in reverse at -1 m/s!"
    \item \textbf{Smooth Driving:} \gls{rmse} of acceleration deviating from 0 $ms^{-2}$, denoted as $E_S [ms^{-2}]$. Instruction prompt: "Reduce jerkyness!"
\end{enumerate}

\begin{mpcbubble}{2a}{Instruction}
Reverse the car!
\end{mpcbubble}

\begin{mpcbubble}{2b}{Prompt}
Adapt the tuneable parameters of the MPC so that the car achieves the following: \texttt{instruction}.\\
This is the MPC formulation: \texttt{base\_memory}.\\
Make use of these memories: \texttt{rag\_memories}.\\
Return format:\\
\texttt{new\_mpc\_params} = \texttt{\{param1: new\_value1, param2: new\_value2, ...\}}
\end{mpcbubble}

\begin{mpcbubble}{2c}{Response}
\texttt{new\_mpc\_params} = \texttt{ \{qv: 0.1, 'qn': 40, qalpha: 50, ddelta\_min: -5,
ddelta\_max: 0, dv\_min: -50, dv\_max: -1, v\_min: -1, v\_max: -1, boundary\_inflation: 0.1\} }\\
\textbf{Explanation:} The new parameters are tuned to achieve the goal of driving the track in reverse at -1.0 m/s. The minimum and maximum velocities (\texttt{v\_min} and \texttt{v\_max}) are set to -1 and -1, respectively, to ensure the velocity is at -1.0 m/s.
\end{mpcbubble}
\setcounter{bubblegroup}{2}

\begin{figure*}[!htb]
    \centering
    \begin{subfigure}[b]{0.49\textwidth}
        \centering
        \includegraphics[height=5cm]{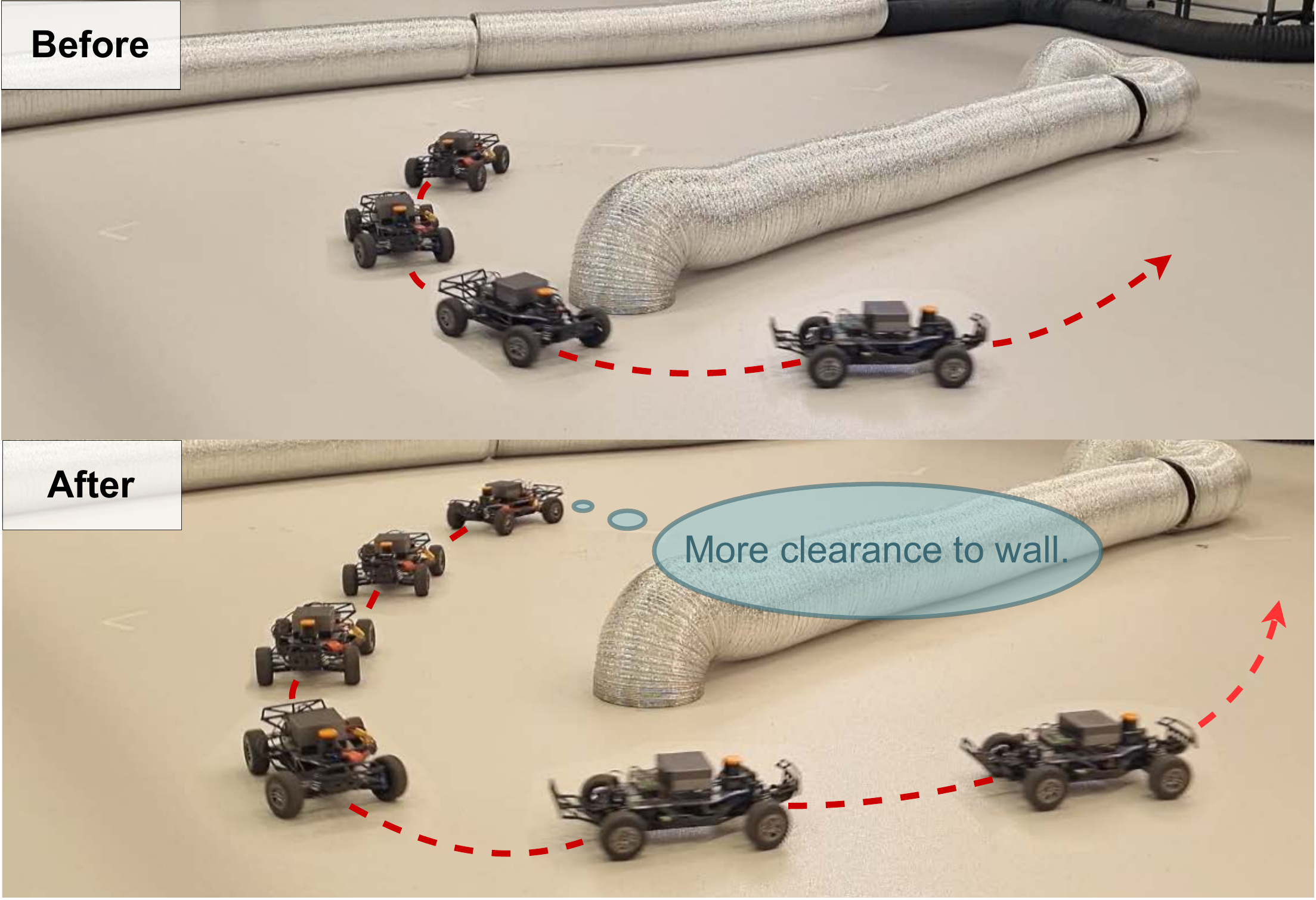}
        \caption{Prompt: \emph{"Drive further away from the wall."}}
        \label{fig:robot_centerline}
    \end{subfigure}
    \hfill
    \begin{subfigure}[b]{0.49\textwidth}
        \centering
        \includegraphics[height=5cm]{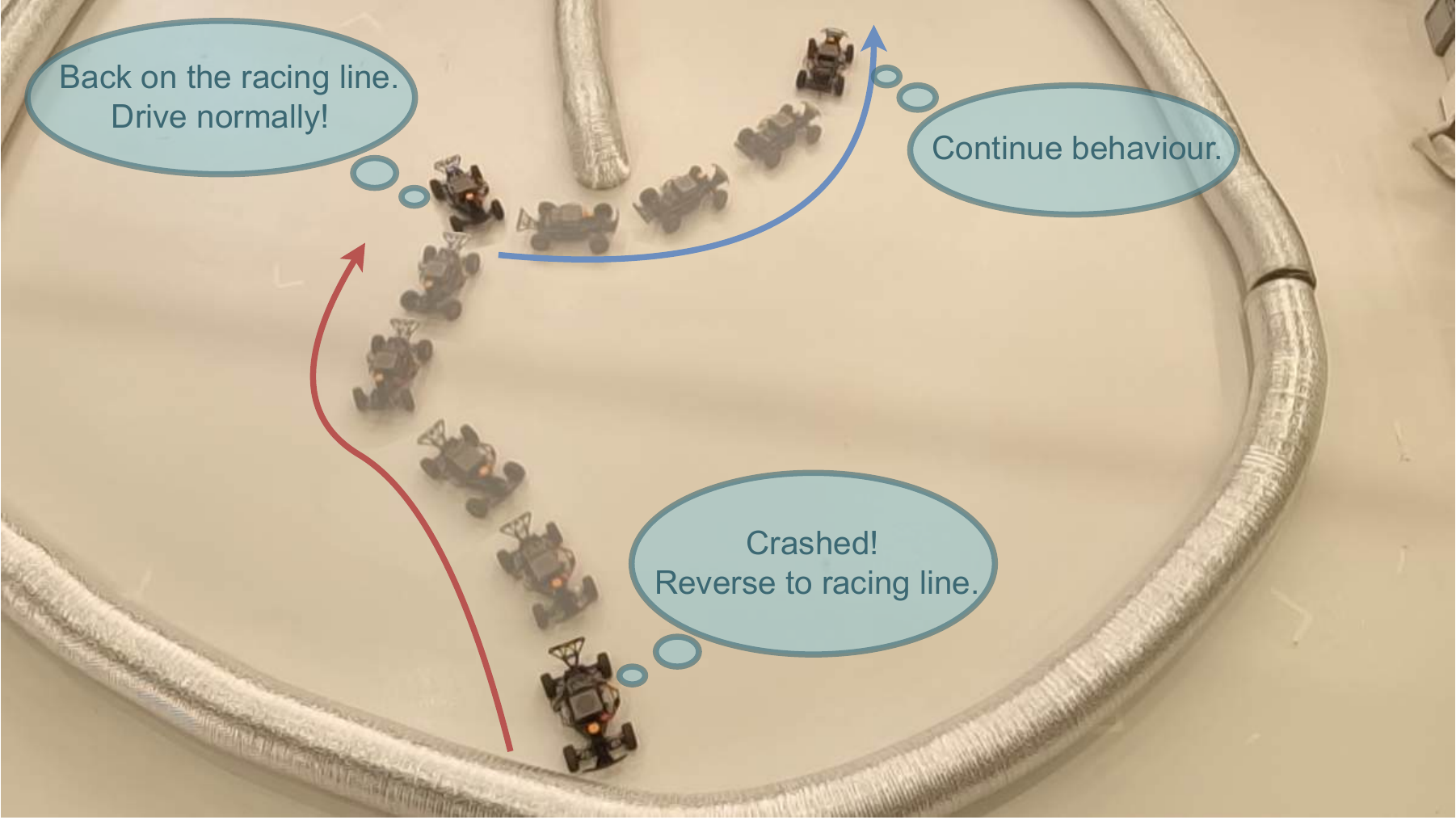}
        \caption{Prompt: \emph{”Drive normally!”}}
        \label{fig:robot_crash_recovery}
    \end{subfigure}
    \caption{Visualization of physical robot behaviors: (a) increasing clearance from walls, (b) recovering from a crash scenario. All experiments were performed utilizing the edge-deployed \textit{Qwen2.5-7b} \gls{llm}.}
    \label{fig:robot_res}
\end{figure*}

It is important to note that while the evaluation scenarios presented here are specifically designed to ensure the computation of quantifiable metrics, such as \gls{rmse}, the \textit{MPCxLLM} module is capable of processing and responding to a wide range of other natural language instructions. The chosen scenarios are simply representative examples where measurable outcomes --- like centerline deviation, velocity tracking, and smooth driving --- allow for clear and reproducible analysis. However, the methodology and interaction flow described are not limited to these examples and can extend to other instructions.

\Cref{tab:control_res} demonstrates the controller adaptability through the evaluated \glspl{llm}. The \gls{mpc} baseline, represents the different \glspl{rmse} adherences with default parameters, tuned for nominal operation of the vehicle. All percentage improvements are reported in absolute percentage points. The results show that the inclusion of \gls{rag} generally improves instruction adherence for locally deployable \glspl{llm} (\textit{Phi3-mini, Qwen2.5}) by approximately 40\%, although performance without \gls{rag} can be negligible or even detrimental. \textit{GPT4o} cannot be deployed via \gls{lora} fine-tuning and is therefore excluded. For locally deployable models, \gls{lora} fine-tuning offers an additional 15–20\% improvement over \gls{rag}-only setups. Hence, in terms of controller adaptability, both \gls{rag} and \gls{lora} demonstrate a 45.1\% and 52.2\%  (for \textit{Phi3-mini} and \textit{Qwen2.5} respectively) improvement over the nominal \gls{mpc} baseline. Lastly, quantization introduces a minor ~1-6\% performance drop but is essential for achieving acceptable computational performance on the \gls{obc}, as detailed in \Cref{subsec:results_compute}. Note that the numerical values should be interpreted with caution, as the closed-loop experiment is not deterministic, since it depends on the \gls{mpc}'s adherence quality and the specific map used in the open-source simulation environment \cite{forzaeth}, hence the relative improvement should be regarded. From the local \glspl{llm}, \Cref{tab:control_res} shows that \textit{Qwen2.5-7b} consistently outperforms \textit{Phi3-mini-3.8b} in terms of controller adaptability.

\Cref{bubble:2a} illustrates a qualitative result of the \textit{MPCxLLM} module. The input instruction is passed to the \gls{llm}, where the prompt demonstrates how the \gls{llm} is guided by the instruction and optionally enriched by the \textit{MPCxLLM} \gls{rag}. Finally, the module's output displays a new set of \gls{mpc} cost and constraint parameters that are subsequently parsed and transmitted to the \gls{mpc} as a \gls{ros} dynamic reconfigurable parameter.

\subsection{Physical Robot}
Multiple qualitative examples of the proposed framework operating on the physical robot are shown in \Cref{fig:robot_res}. In \Cref{fig:robot_centerline}, the human instructs the robot to increase its distance from the wall. The \emph{before} image illustrates how close the robot was initially driving to the wall, while the \emph{after} image demonstrates a much safer clearance achieved through adjustments made by the \textit{MPCxLLM} module. The \textit{MPCxLLM} was prompted with: \textit{"Drive further away from the wall."}

\Cref{fig:robot_crash_recovery} showcases the \textit{DecisionxLLM} and \textit{MPCxLLM} modules combined, by detecting a crash and subsequently instructing the robot to reverse and then safely resume its path. The \textit{DecisionxLLM} module is instructed by the human to: \textit{"Drive normally!"}. The full model outputs of the \gls{llm} is shown in \Cref{bubble:3a}. More qualitative experiments on the physical robot are depicted in \Cref{appendix:errata}.

\subsection{Computation} \label{subsec:results_compute}
\begin{table}[!htb] 
\small
    \centering 
    \setlength{\tabcolsep}{0.7 mm}
    \begin{tabular}{l|l|c|c|c|c|c|cc}
    \toprule
    \textbf{HW} & \textbf{LLM} & \textbf{Quant} & \textbf{Param} & \textbf{Mem} & \textbf{Tokens} & \textbf{Tokens/s} & \multicolumn{2}{c}{\textbf{Latency [s]}$\downarrow$} \\
    & & & \textbf{[\#B]} & \textbf{[GB]} & \textbf{[\#]} & \textbf{[s$^{-1}$]$\uparrow$} & $\mu_{t}$ & $\sigma_{t}$ \\
    \midrule
    RTX & Phi3 & FP16 & 3.8 & 4.3 & 72 & \textbf{25.23} & \textbf{2.85} & \textbf{0.05} \\
    3090 & Qwen  & FP16 & 7 & 7.8 & 50 & 11.14 & 4.51 & 0.08 \\
    \cmidrule{2-9}
    & Phi3 & Q5 & 3.8 & 3.9 & 110 & \textbf{148.36} & \textbf{0.75} & \textbf{0.06} \\
    & Qwen  & Q5 & 7 & 5.6 & 97 & 107.52 & 0.91 & 0.08 \\
    \midrule
    \multirow{3}{*}{\tabincell{l}{Jetson\\Orin\\AGX}} & Phi3 & FP16 & 3.8 & 4.2 & 72 & \textbf{4.71} & \textbf{15.29} & \textbf{0.48} \\
    & Qwen  & FP16 & 7 & 6.9 & 64 & 2.11 & 30.48 & 0.59 \\
    \cmidrule{2-9}
    & Phi3 & Q5 & 3.8 & 3.6 & 154 & \textbf{32.47} & \textbf{4.80} & \textbf{0.58} \\
    & Qwen  & Q5 & 7 & 5.3 & 121 & 22.12 & 5.52 & \textbf{0.58} \\
    \bottomrule
    \end{tabular}
    \caption{Comparison of computational performance for locally deployable models, \textit{Phi3-mini-3.8b} and \textit{Qwen2.5-7b}. The \glspl{llm} were deployed on both an RTX 3090 GPU and the \textit{Jetson Orin AGX} robotic \gls{obc}. FP16 denotes full-precision models, while Q5 represents the \texttt{Q5\_k\_m} quantized models implemented via the \texttt{llama.cpp} inference engine. The number of tokens denotes the output tokens generated for the given inference. The average inference latency with the standard deviation is denoted with $\mu_{t}, \sigma_{t}$ respectively.}
    \label{tab:compute}
\end{table}

Given the need for efficient hardware utilization in autonomous driving scenarios and the necessity for real-time interaction, the efficiency based on two models is evaluated and discussed in \Cref{tab:compute} for the locally deployable \textit{Phi3-mini-3.8b} and \textit{Qwen2.5-7b}. The same input prompt was used for all compute evaluations and performed 60 times sequentially.

For the framework based on \textit{Phi3}, when deployed on RTX 3090 hardware using FP16, the model with 3.8 billion parameters achieves a token output rate of 25.23 tokens per second and utilizes 4.3 GB of memory. In contrast, when quantized to \texttt{Q5\_k\_m}, the memory usage decreases to 3.9 GB, and the throughput speed significantly increases to 148.36 tokens per second. On the computationally constrained \textit{Jetson Orin} hardware, the FP16 Phi3 achieves an inference time of 15.29 seconds, while the \texttt{Q5\_k\_m} quantized model shows improved performance with an inference time of 4.80 seconds for 154 tokens.
For the framework based on Qwen, the outcome is similar. On the RTX 3090, when the model is quantized to \texttt{Q5\_k\_m}, the throughput speed rises to nearly 9.7$\times$ that of FP16, reaching 107.52 tokens per second. Thus, the inference time decreases significantly to 0.91 seconds, approximately 20\% of the time required for FP16. On the \textit{Jetson Orin}, the \texttt{Q5\_k\_m} configuration increases the throughput to about 10.5$\times$ compared to the FP16, at 22.12 tokens per second. The inference time is lowered to 5.52 seconds, less than 20\% of that of the FP16. These results highlight the substantial efficiency gains achievable through post-training quantization, enabling the deployment of our \glspl{llm}-based framework in computationally constrained hardware while maintaining real-time capabilities on robotic platforms.

\section{Conclusion} \label{sec:conclusion}

This work introduces a hybrid architecture that enables the integration of low-level \gls{mpc} and edge-deployed \glspl{llm} to enhance robotic decision-making and \gls{hmi} through natural language in \gls{ads}. It offers the flexibility to choose between different \glspl{llm} based on operational requirements such as cloud connectivity, privacy considerations, and latency constraints. This approach bridges the gap between high-level reasoning and low-level control adaptability.

On locally deployed \glspl{llm}, the \textit{DecisionxLLM} module demonstrates up to 10.45\% improvement in reasoning accuracy when augmented with \gls{rag} and \gls{lora} fine-tuning. The \textit{MPCxLLM} module showcases controller adaptability, achieving up to a 52\% improvement in controller adaptability, highlighting how natural language can adjust low-level \gls{mpc} parameters to achieve flexible robotic behaviors while maintaining safety and constraint satisfaction through \gls{mpc} systems.

Furthermore, this work demonstrates the deployment of embodied \gls{ai} locally on embedded platforms, highlighting the importance of quantization in enabling real-time performance. Through post-training quantization, a 10.5$\times$ improvement in throughput are achieved for the \textit{Qwen2.5-7b} model on the \textit{Jetson Orin AGX} \gls{obc}, allowing for efficient deployment of \glspl{llm} on resource-constrained hardware.


\section{Limitations} \label{sec:limitations}
One limitation of this approach is the relatively slow decision-making and controller adaptation process, which may fail to capture subtle (or high-frequency) behavioral nuances in the robotic state. Additionally, the reliance on text-based \glspl{llm} constrains the reasoning capabilities to state-based information alone. In contrast, multimodal \glspl{llm} could significantly enhance performance by incorporating visual data, enabling richer and more context-aware reasoning. However, the computational constraints of the embedded \gls{obc} necessitated the use of standard \glspl{llm} as an initial step. Future work may address these constraints by exploring efficient deployment strategies for multimodal models on resource-limited hardware and investigating their reasoning capabilities. Lastly, locally deployed \glspl{llm} are not without flaws. They occasionally introduce reasoning errors and inconsistencies in controller adaptability, highlighting areas for improvement by retraining with a larger amount of distillation data. The proposed framework should therefore be viewed as a potential approach to integrating knowledge into \gls{ads}.

\section*{Acknowledgments}
This work is funded in part by the dAIEDGE project supported by the EU Horizon Europe research and innovation programme under Grant Agreement Number: 101120726.


\bibliographystyle{plainnat}
\bibliography{references}

\appendices
\onecolumn
\section{LoRA Fine-Tuning Examples}\label[appendix]{appendix:lora}
For \textit{DecisionxLLM}, we generate synthetic data via a two-stage process. First, in simulation, the vehicle collects diverse states (e.g., centerline tracking, reversing, unsafe maneuvers). These states are input to \textit{GPT4o}, with prompts enriched by a \gls{rag} context (\Cref{lst:decision_rag}) that injects robot-specific details (e.g., speed ranges, safe distances). Parameter values are randomized programmatically to avoid overfitting, ensuring diversity in the synthetic dataset. \textit{GPT4o} then produces behavior descriptions based on queries such as: \textit{"The human wants to: \texttt{Drive Safely}. Check if this state: \texttt{robot\_state} adheres to the command. Additional context: \texttt{RAG\_info}. Decide: (a) Yes, (b) No, adjust behavior."} This yields 626 state-prompt response pairs for \textit{DecisionxLLM} fine-tuning.

For \textit{MPCxLLM}, \textit{GPT4o} is prompted with the human \texttt{instruction} (e.g., \textit{"Do not exceed speeds of 10 km/h."}), \texttt{base\_memory} (\Cref{lst:base_mpc}), and \texttt{RAG\_memories} (\Cref{lst:mpc_rag}). Programmatic randomization of the numeric inputs prevents overfitting, producing 150 prompt response pairs for LoRA fine-tuning of MPC adaptation.

\section{Additional MPC Information} \label[appendix]{appendix:mpc}
The baseline \gls{mpc} parameters within \Cref{tab:control_res} have been set to ensure nominal tracking of the racing line in the simulation environment, which are the default parameters visible in \Cref{lst:base_mpc}. Hence, in the controller adaptability experiment \Cref{subsec:results_mpcxllm}, the \textit{MPCxLLM} module was tasked to adapt the nominal behavior (e.g., from tracking the racing line to \textit{"Reverse the car!"}) as described in \Cref{subsec:results_mpcxllm} --- thus yielding a quantitative and measurable alteration of the nominal (baseline) \gls{mpc} behavior.\newline

The base memory of \Cref{lst:base_mpc} within the \textit{MPCxLLM} prompt \Cref{bubble:2a} serves as a predefined knowledge foundation for mapping high-level natural language instructions into precise \gls{mpc} parameters. It encodes the cost structure, tuneable parameters, and safety constraints essential for dynamic control adjustments. Each parameter is strictly defined with names, valid ranges, and default values that are defined as \gls{ros} dynamic reconfigure parameters.

\begin{lstlisting}[caption={MPCxLLM Base Memory}, label={lst:base_mpc}]
# Cost expression with adjustable weights:
model.cost_expr_ext_cost = (
    weight_qn * n**2 + 
    weight_qalpha * alpha**2 + 
    weight_qv * (v - V_target)**2 + 
    weight_qac * der_v**2 + 
    weight_ddelta * derDelta**2 +
    u.T @ R @ u
)

# Tuneable cost weights and constraints (USE EXACT NAMES, DO NOT CREATE NEW ONES!):
# param: min, max, default # description
qv 0, 2, 10 # Velocity weight: minimizes speed tracking error
qn 0, 100, 20 # Lateral weight: minimizes deviation from the track
qalpha 0, 100, 7 # Heading weight: minimizes orientation error
qac 0, 1, 0.01 # Acceleration weight: penalizes high acceleration
qddelta 0, 100, 0.1 # Steering weight: penalizes fast steering changes
alat_max 0, 20, 10 # Max lateral acceleration: limits side force
a_min -20, 0, -5 # Min acceleration: lower acceleration bound
a_max 0, 20, 5 # Max acceleration: upper acceleration bound
v_min -2, 5, 1 # Min velocity: lower speed bound
v_max -1, 10, 5 # Max velocity: upper speed bound
track_safety_margin 0, 1.0, 0.45 # Safety margin: increases track boundary margin
\end{lstlisting}

\Cref{lst:mpc_rag} shows the \textit{MPCxLLM} \gls{rag} memories establish context-specific mappings between natural language instructions and corresponding \gls{mpc} parameter adjustments. Each memory entry provides guidance on how parameters from the \texttt{base\_memory} influence the \gls{mpc}, resembling simplified \emph{driving school} instructions, akin to \cite{wen2023dilu}. These memory entries are modular and can be combined to address more complex scenarios. The \gls{rag} mechanism ensures effective retrieval by performing similarity matching between the user instruction and scenario descriptions.

\begin{lstlisting}[caption={MPCxLLM RAG Memories}, label={lst:mpc_rag}]
# Memory Entry 0:
Scenario:
To force going forwards v_min should be positive. If you want it to be able to reverse, then set v_min to negative.
MPC Action:
mpc_params = {
    'v_min': positive, if you want to go forwards, else negative to reverse
}

# Memory Entry 1:
Scenario:
Always have v_max be higher than v_min.
MPC Action:
mpc_params = {
    'v_max': higher than v_min
}

...

# Memory Entry 10:
Scenario:
To minimize the lateral acceleration and jerk, set alat_max to a low value and a_min and a_max close to zero. If you want to drive more aggressive, then set alat_max to a higher value.
MPC Action:
mpc_params = {
    'alat_max': low for minimizing lateral acceleration, else high for aggressive driving
}
\end{lstlisting}

\section{Additional Decision-Making Information} \label[appendix]{appendix:decision}
Decision accuracy was evaluated using a dataset of 200 state samples (using a different state-dataset as the \gls{lora} fine-tuning data), yielding 1600 state-command pairs across 8 driving commands \Cref{subsec:results_decisionxllm}. Each state corresponds to a command (e.g., \textit{"Drive faster than 3 m/s."}). The \textit{DecisionxLLM} module performs binary classification to determine if the robot’s state meets the command (\Cref{fig:graphical_abstract} or \Cref{fig:decision_llm}). Adherence is verified computationally; for instance, for \textit{"Reverse the car!"}, the system checks if the velocity is negative. Accuracy is measured as the percentage of correctly classified state-command pairs.\newline

\Cref{lst:decision_rag} is an excerpt is from the \textit{DecisionxLLM} \gls{rag} entries. The \gls{rag} enhances the \texttt{human\_prompt} by augmenting it with context retrieved through similarity searches across various hints. The top-k most relevant hints are selected to enrich the prompt, providing additional contextual grounding. This approach allows users to ground the robot using natural language. For instance, a user might specify a different desired nominal speed range, such as operating the car between 2-\SI{4}{\metre\per\second}, thereby supplying \textit{DecisionxLLM} with updated operational constraints.

\begin{lstlisting}[caption={DecisionxLLM RAG Memories}, label={lst:decision_rag}]
# Hint 1:
If the d-speed is above than 0.5m/s is high.

# Hint 2:
Unless specified differently by the human, the car is usually driving at speeds between 5 and 7m/s.

# Hint 3:
If the distance to a wall is smaller than 0.4m, the car is close to that wall.

...

# Hint 8:
A d-coordinate above 0.3m is considered not to be on the racing line.

# Hint 9:
The car is oscillating if the d-coordinate oscillates between positive and negative values exceeding a magnitude of 0.3 metres.

# Hint 10:
Oscillations in d-coordinate under 0.3m or d-speed under 0.5m/s are negligible.
\end{lstlisting}

\section{Crash Recovery Output Examples on the Physical Robot} \label[appendix]{appendix:errata} 
\Cref{bubble:3a} captures the chat log from the experiment detailed in \Cref{fig:robot_crash_recovery}. The robot begins in a crashed state, facing a wall. The \textit{DecisionxLLM} identifies this condition and adjusts the \gls{mpc} to initiate a reversing maneuver. Once the robot successfully reverses out of the crash, it resumes normal driving along the track. This log presents the \gls{llm} output generated during the experiment, where the locally deployed \textit{Qwen2.5-7b} was running on the \gls{obc}.

\begin{decisionbubble}{3a}{Instruction}
Drive normally!
\end{decisionbubble}

\begin{decisionbubble}{3b}{Prompt}
The human wants to: \textit{"Drive normally!"}\\
The data has been sampled for 2.0 seconds in 5 samples.
\begin{itemize}
    \item \textbf{s-speed}: \texttt{-0.009, -0.026, -0.03, -0.03, -0.013} 
    \item \textbf{d-speed}: \texttt{0.096, 0.091, 0.088, 0.09, 0.096} 
    \item \textbf{distance to left wall}: \texttt{2.953, 2.954, 2.954, 2.955, 2.955}
    \item \textbf{distance to right wall}: \texttt{0.149, 0.148, 0.148, 0.147, 0.147}
    \item \textbf{crashed:} \texttt{True}
\end{itemize}
Here are some guides to help you reason: \texttt{rag\_hints}.\\
Check if the car is doing what the human wants. Choose one of the following actions to command the car: 
\begin{itemize}
    \item a) Continue behavior
    \item b) Change behavior: \texttt{instruction}
\end{itemize}
\end{decisionbubble}

\begin{decisionbubble}{3c}{Response}
The car is \textbf{facing a crash} and is reversing, which is \textbf{not the normal driving behavior}. The car should be facing forward and driving at a normal speed. The s-speed is close to 0m/s, indicating the car is stopped, and the d-speed is high, suggesting the car is oscillating around the racing line. The car is also close to the right wall, which is a sign of a crash. \textbf{Reversing is necessary} to get the car back on the racing line and back to normal driving.

\texttt{Action:} \\
\textbf{Change behavior:} \texttt{Reverse the car to get back on the racing line.}
\end{decisionbubble}

\begin{mpcbubble}{3d}{Response}
\texttt{new\_mpc\_params} = \texttt{\{qv: 0.5, qn: 50.0, qalpha: 5.0, qac: 0.01, qddelta: 0.1, alat\_max: 10.0, a\_min: -5.0, a\_max: 5.0, \textbf{v\_min: -2.0, v\_max: -1.0}, track\_safety\_margin: 0.45\}}\\
\textbf{Explanation:} Setting parameters to reverse the car.
\end{mpcbubble}

\begin{narratebubble}
The car was stuck in the wall. The \textit{DecisionxLLM} stage made the correct decision, albeit a reasoning error on stating that the d-speed is high, but this did not change the overall decision-making. Then the \textit{MPCxLLM} stage correctly adapts the \gls{mpc} to induce reversing behavior.
\end{narratebubble}

\begin{decisionbubble}{3e}{Prompt}
The human wants to: \textit{"Drive normally!"}\\
The data has been sampled for 2.0 seconds in 5 samples.
\begin{itemize}
    \item \textbf{s-speed}: \texttt{-0.022, -0.014, -0.037, -0.005, -0.038} 
    \item \textbf{d-speed}: \texttt{0.091, 0.095, 0.088, 0.097, 0.088} 
    \item \textbf{distance to left wall}: \texttt{2.956, 2.955, 2.955, 2.955, 2.956}
    \item \textbf{distance to right wall}: \texttt{0.146, 0.147, 0.147, 0.147, 0.146}
    \item \textbf{crashed:} \texttt{True}
\end{itemize}
Here are some guides to help you reason: \texttt{rag\_hints}.\\
Check if the car is doing what it needs to ensure recovery. Choose one of the following actions to command the car:
\begin{itemize}
    \item a) Continue behavior
    \item b) Change behavior: \texttt{instruction}
\end{itemize}
\end{decisionbubble}

\begin{decisionbubble}{3f}{Response}
The \textbf{car has crashed and is reversing, so it should continue to reverse} until it is safe to proceed. The car's speed is very low, indicating it is stopped, and it is close to the right wall. Therefore, the car should reverse slowly to get back on the racing line.

\texttt{Action:} \\
\textbf{Continue behavior}
\end{decisionbubble}

\begin{narratebubble}
Even though reversing behavior was already induced, the state was sampled while the car was still crashed into the wall. Hence the decision of continuing the reversing behavior is correct.
\end{narratebubble}

\begin{decisionbubble}{3g}{Prompt}
The human wants to: \textit{"Drive normally!"}\\
The data has been sampled for 2.0 seconds in 5 samples.
\begin{itemize}
    \item \textbf{s-speed}: \texttt{-0.254, -0.149, -0.086, -0.088, -0.029} 
    \item \textbf{d-speed}: \texttt{0.748, 0.281, 0.163, 0.155, 0.112} 
    \item \textbf{distance to left wall}: \texttt{0.372, 0.379, 0.386, 0.393, 0.391}
    \item \textbf{distance to right wall}: \texttt{2.82, 2.846, 2.839, 2.832, 2.834}
    \item \textbf{crashed:} \texttt{False}
\end{itemize}
Here are some guides to help you reason: \texttt{rag\_hints}.\\
Check if the car is doing what the human wants. Choose one of the following actions to command the car: 
\begin{itemize}
    \item a) Continue behavior
    \item b) Change behavior: \texttt{instruction}
\end{itemize}
\end{decisionbubble}

\begin{decisionbubble}{3h}{Response}
The car is reversing, which is \textbf{not the normal driving behavior}. The car should be facing forward to maintain normal driving. Therefore, the action should be to \textbf{stop the car from reversing and resume normal driving}.\\
\texttt{Action:} \\
\textbf{Change behavior:} \texttt{Stop reversing and resume normal driving.}
\end{decisionbubble}

\begin{mpcbubble}{3j}{Response}
\texttt{new\_mpc\_params} = \texttt{\{'v\_min': 1.5, 'v\_max': 1.5, 'qn': 100, 'qalpha': 100, 'qac': 0.01, 'qddelta': 0.1, 'alat\_max': 10, 'a\_min': -5, 'a\_max': 5\}}\\
\textbf{Explanation:} The parameters are tuned to smoothly resume normal driving.
\end{mpcbubble}

\begin{narratebubble}
It is correctly detected that the car is no longer crashed and should now drive forwards again. The \gls{mpc} parameters are then again adjusted to do so.
\end{narratebubble}

\setcounter{bubblegroup}{3}

\subsection{Velocity Adherence on the Physical Robot}
\begin{figure}[!htb]
    \centering
    \includegraphics[scale=0.4]{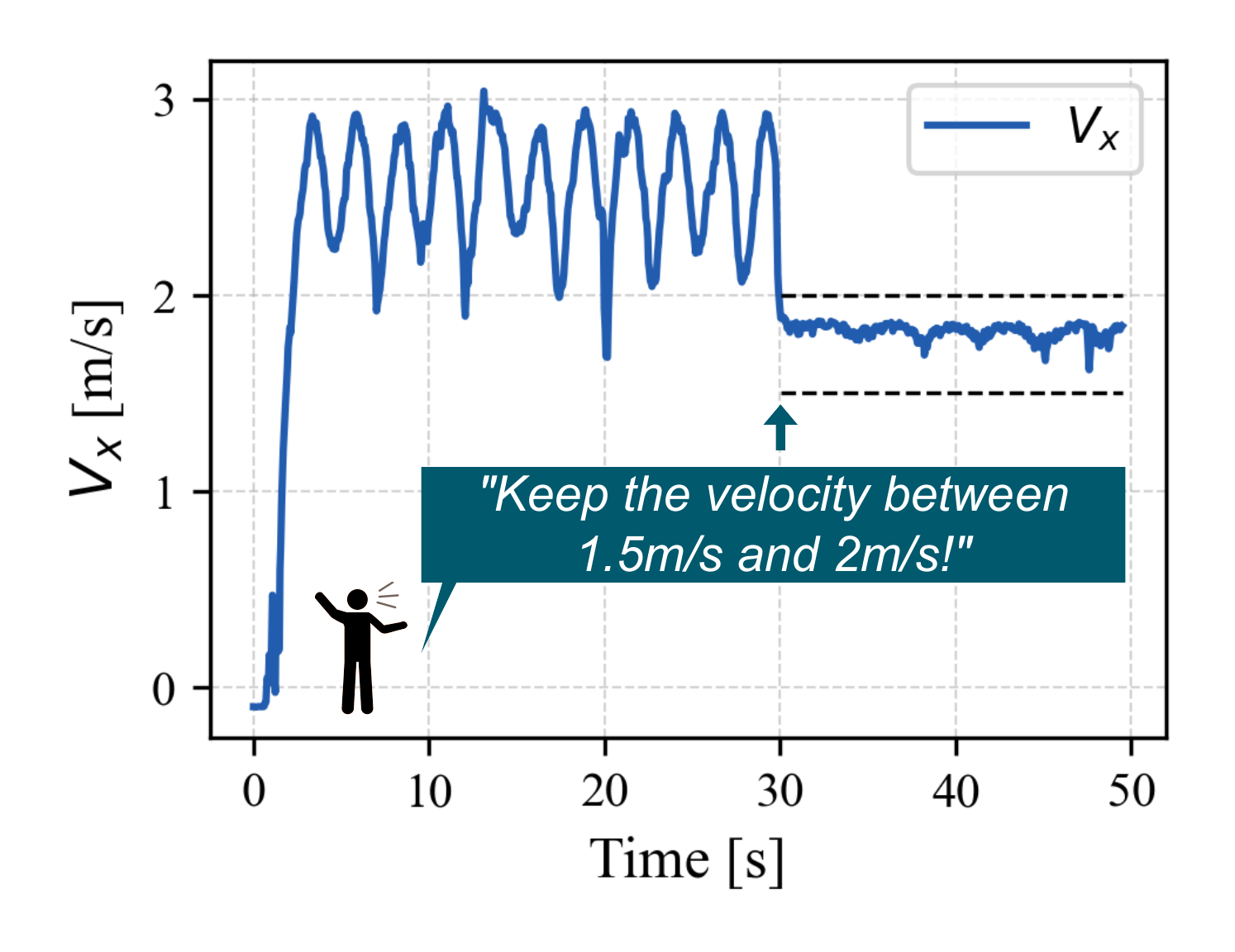}
    \caption{Illustration of the robot's velocity behavior before and after (at 30s) the human prompts the \gls{llm}: \textit{"Drive at speeds between 1.5 and 2.0 m/s"}.}
    \label{fig:robot_vel}
\end{figure}

\label[appendix]{appendix:veladherence}
\cref{fig:robot_vel} illustrates the robot's ability to respond to human commands effectively by adjusting its velocity to remain within a specific range. This experiment is performed on the physicial robot using the edge-deployed \textit{Qwen2.5} model. Upon receiving the instruction to maintain a target velocity window, the robot promptly modifies its speed, demonstrating precise control and adaptability. This behavior highlights the system's capability to monitor and regulate its performance, ensuring that the velocity consistently stays within the designated limits.

\section{Extended Decision-Making Results}\label[appendix]{appendix:extended_res}

\Cref{tab:ext_decision_res} extends \Cref{tab:decision_res} by presenting the performance of various \glspl{llm} across individual decision-making test cases, with the average accuracy reported in the final column. Additionally, this table includes evaluations of contemporary models such as \emph{Phi4-14b}\footnote{\url{https://huggingface.co/unsloth/phi-4}}, \emph{Gemma2-9b}\footnote{\url{https://huggingface.co/unsloth/gemma-2-9b}}, and \emph{DeepSeek R1-distilled Qwen2.5-7b}\footnote{\url{https://huggingface.co/unsloth/DeepSeek-R1-Distill-Qwen-7B}}. These models represent state-of-the-art architectures at the time of this submission. However, due to time constraints, the quantization and \texttt{Q5\_k\_m GGUF} inference pipeline could not be implemented.

Furthermore, the large parameter count (or slower inference architecture) of \emph{Phi4-14b} and \emph{Gemma2-9b} --- and in the case of \emph{Qwen2.5-7b-R1distill}, the extended \gls{cot} reasoning process --- renders them infeasible for closed-loop \gls{mpc} and simulation experiments. As a result, only decision-making results are presented here.

A key observation from the results is that, with and without \gls{rag}, \emph{Qwen2.5-7b-R1distill} demonstrates a significant performance boost over its non-R1-distilled counterpart of the same parameter count. The $\dagger$ symbol denotes instances where the \gls{llm} was occasionally cut off during evaluation --- despite doubling the maximum token limit to 1024 --- directly leading to incorrect decision scores. This issue stems from the computational overhead associated with R1 \gls{cot} reasoning, resulting in a significantly higher number of output tokens and consequently longer inference and evaluation times. This means, that given more time, the performance values would probably be even higher.

Moreover, \emph{Qwen2.5-7b-R1distill} exhibits a notable drop in performance when fine-tuned with \gls{lora}. This finding is particularly intriguing, as the \gls{lora} fine-tuning was based on \gls{sft} using \emph{GPT-4o} data, suggesting a potential degradation in reasoning capability rather than improvement and that the \gls{lora} tuning step could perhaps be improved by utilizing R1 generated data for \gls{sft}. Further, it is to note, that once \gls{lora} tuned on \emph{GPT4o} data, the \gls{llm} now again learns to be concise and is no longer cut-off, however this comes at the cost of reasoning accuracy performance.

\begin{table}[!htb]
    \centering
    \begin{adjustbox}{max width=\columnwidth}
    \begin{tabular}{l|c|c|c|c|c|c|c|c|c|c|c|c|c}
    \toprule
    \textbf{Model} & \textbf{Params} & \textbf{Quant} & \textbf{LoRA} & \textbf{RAG} & \textbf{Centerline} & \textbf{Close Wall} & \textbf{Forward} & \textbf{Oscillating} & \textbf{Racingline} & \textbf{Reversed} & \textbf{Speed} & \textbf{Stop} & \textbf{Avg. Accuracy} \\
    \midrule
    GPT4o & ? & \xmark & \xmark & \xmark & \textbf{94.57} & 80.5 & \textbf{81} & 80.5 & \textbf{81} & 70.83 & 77.5 & \textbf{87.5} & \textbf{81.68} \\
    Phi3-mini & 3.8B & \xmark & \xmark & \xmark & 83.7 & 56.5 & 76 & 76.5 & 74 & 48.96 & 74.5 & 87 & 72.15 \\
    Phi4-14b & 14B & \xmark & \xmark & \xmark & 91.3 & \textbf{83.5} & 74.5 & 75.5 & 71 & 62.5 & 71 & 77 & 75.79 \\
    Gemma2-9b & 9B & \xmark & \xmark & \xmark & 84.78 & 86 & 80 & \textbf{81.5} & 78 & \textbf{84.9} & \textbf{77.5} & 83.5 & 82.02 \\
    Qwen2.5-7b & 7B & \xmark & \xmark & \xmark & 82.07 & 71 & 76 & 84 & 75.5 & 72.4 & 71.5 & 89.5 & 77.75 \\
    Qwen2.5-7b-R1distil & 7B & \xmark & \xmark & \xmark & 92.93 & 84 & 75 & 79 & 80 & 72.4 & 74.5 & 82.5 & 80.04$\dagger$ \\
    \midrule
    GPT4o & ? & \xmark & \xmark & \cmark & \textbf{100} & \textbf{98} & \textbf{97} & \textbf{85.5} & \textbf{95.5} & \textbf{83.85} & \textbf{96} & 84 & \textbf{92.48} \\
    Phi3-mini & 3.8B & \xmark & \xmark & \cmark & 90.76 & 66 & 81 & 84 & 78 & 68.23 & 77 & 84.5 & 78.69 \\
    Phi4-14b & 14B & \xmark & \xmark & \cmark & 97.83 & 84 & 86.5 & 75 & 85 & 65.62 & 77 & 85 & 81.99 \\
    Gemma2-9b & 9B & \xmark & \xmark & \cmark & 96.2 & 82 & 79.5 & 83 & 79 & 71.35 & 74 & \textbf{85} & 81.26 \\
    Qwen2.5-7b & 7B & \xmark & \xmark & \cmark & 90.79 & 78.5 & 83 & 84 & 81 & 73.44 & 78.5 & 90.5 & 82.47 \\
    Qwen2.5-7b-R1distil & 7B & \xmark & \xmark & \cmark & 94.02 & 89 & 88.5 & 82.5 & 86.5 & 78.12 & 79 & 86.5 & 85.52$\dagger$ \\
    \midrule
    Phi3-mini & 3.8B & \xmark & \cmark & \cmark & 94.57 & 68.5 & 82.5 & 80.5 & 82 & 79.69 & 75.5 & \textbf{97.5} & 82.60 \\
    Phi4-14b & 14B & \xmark & \cmark & \cmark & \textbf{100} & \textbf{90.5} & \textbf{94} & 80.5 & \textbf{93.5} & 79.17 & \textbf{91.5} & 89.5 & \textbf{89.83} \\
    Gemma2-9b & 9B & \xmark & \cmark & \cmark & 82.61 & 72 & 76 & 75 & 73 & 72.4 & 66.5 & 72 & 73.69 \\
    Qwen2.5-7b & 7B & \xmark & \cmark & \cmark & 97.28 & 78.5 & 90 & \textbf{86} & 88.5 & \textbf{81.77} & 81 & 95.5 & 87.32 \\
    Qwen2.5-7b-R1distil & 7B & \xmark & \cmark & \cmark & 85.33 & 73 & 81.5 & 79.5 & 85.5 & 85.42 & 83 & 93 & 83.28 \\
    \midrule
    Phi3-mini & 3.8B & \cmark & \cmark & \cmark & 88.59 & 84.5 & 84 & 74.5 & 90 & 86.98 & 85 & 86 & 84.95 \\
    Qwen2.5-7b & 7B & \cmark & \cmark & \cmark & \textbf{98.91} & \textbf{81.5} & \textbf{85} & \textbf{87.5} & \textbf{85.5} & 80.21 & \textbf{82} & \textbf{95.5} & \textbf{87.02} \\
    \bottomrule
    \end{tabular}
    \end{adjustbox}
    \caption{Extended performance evaluation of \Cref{tab:decision_res} showcasing the single decision making test cases and additional \glspl{llm}. Accuracy units are reported in \%. ? denotes proprietary information. $\dagger$ denotes that the evaluation was occasionally clipped to the max tokens set to 1024 per inference, which might indicate that a higher score could potentially have been achieved with more time.}
    \label{tab:ext_decision_res}
\end{table}

\end{document}

%% file: acr.tex
\newacronym{llm}{LLM}{Large Language Model}
\newacronym{vlm}{VLM}{Vision Language Model}
\newacronym{vla}{VLA}{Vision Language Action}
\newacronym{rag}{RAG}{Retrieval Augmented Generation}
\newacronym{lora}{LoRA}{Low Rank Adaptation}
\newacronym{mpc}{MPC}{Model Predictive Controller}
\newacronym{rmse}{RMSE}{Root Mean Square Error}
\newacronym{hmi}{HMI}{Human Machine Interaction}
\newacronym{ads}{ADS}{Autonomous Driving Systems}
\newacronym{ml}{ML}{Machine Learning}
\newacronym{nn}{NN}{Neural Network}
\newacronym{rl}{RL}{Reinforcement Learning}
\newacronym{ai}{AI}{Artificial Intelligence}
\newacronym{obc}{OBC}{OnBoard Computer}
\newacronym{gpu}{GPU}{Graphics Processing Unit}
\newacronym{cpu}{CPU}{Central Processing Unit}
\newacronym{ram}{RAM}{Random Access Memory}
\newacronym{ros}{ROS}{Robot Operating System}
\newacronym{peft}{PEFT}{Parameter Efficient Fine-Tuning}
\newacronym{cot}{CoT}{Chain of Thought}
\newacronym{sft}{SFT}{Supervised Fine-Tuning}

%% file: bubbles.tex
\newcounter{bubblegroup}

\crefname{lstlisting}{Listing}{Listings}
\Crefname{lstlisting}{Listing}{Listings}
\crefname{bubblegroup}{Bubble}{Bubbles}
\Crefname{bubblegroup}{Bubble}{Bubbles}

\definecolor{mpcinstructcolor}{HTML}{EFD0E3}
\definecolor{mpcpromptcolor}{HTML}{DC9EC9}
\definecolor{mpcoutcolor}{HTML}{CA6CAE}
\definecolor{usercolor}{HTML}{CCE4EA}
\definecolor{robocolor}{HTML}{99CAD5}
\definecolor{llmcolor}{HTML}{66AFC0}

\newtcolorbox{narratebubble}{%
  colback=gray!20, 
  colframe=gray, 
  fonttitle=\bfseries,
  sharp corners=south,
  boxrule=0.5mm,
  leftrule=0.5mm,
  rightrule=0.5mm,
  toprule=0.5mm,
  bottomrule=0.5mm,
  title=Human Narration,
  coltitle=white,
  enhanced,
  breakable,
}

\newtcolorbox[use counter=bubblegroup]{mpcbubble}[2]{%
  colback=mpcpromptcolor!20,
  colframe=mpcpromptcolor,
  fonttitle=\bfseries,
  sharp corners=south,
  boxrule=0.5mm,
  leftrule=0.5mm,
  rightrule=0.5mm,
  toprule=0.5mm,
  bottomrule=0.5mm,
  label type = bubblegroup,
  label={bubble:#1}, 
  title=Bubble~#1: MPCxLLM -- #2,
  coltitle=white,
  enhanced,
  breakable,
}

\newtcolorbox[use counter=bubblegroup]{decisionbubble}[2]{%
  colback=robocolor!20,
  colframe=robocolor,
  fonttitle=\bfseries,
  sharp corners=south,
  boxrule=0.5mm,
  leftrule=0.5mm,
  rightrule=0.5mm,
  toprule=0.5mm,
  bottomrule=0.5mm,
  label type = bubblegroup,
  label={bubble:#1}, 
  title=Bubble~#1: DecisionxLLM -- #2,
  coltitle=white,
  enhanced,
  breakable,
}

\newtcolorbox[use counter=bubblegroup]{whichllmbubble}[3]{%
  colback=robocolor!20,
  colframe=robocolor,
  fonttitle=\bfseries,
  sharp corners=south,
  boxrule=0.5mm,
  leftrule=0.5mm,
  rightrule=0.5mm,
  toprule=0.5mm,
  bottomrule=0.5mm,
  label type = bubblegroup,
  label={bubble:#1}, 
  title=Bubble~#1: DecisionxLLM -- #2 -- #3,
  coltitle=white,
  enhanced,
  breakable,
}

%% file: main.bbl
\begin{thebibliography}{32}
\providecommand{\natexlab}[1]{#1}
\providecommand{\url}[1]{\texttt{#1}}
\expandafter\ifx\csname urlstyle\endcsname\relax
  \providecommand{\doi}[1]{doi: #1}\else
  \providecommand{\doi}{doi: \begingroup \urlstyle{rm}\Url}\fi

\bibitem[Abdin et~al.(2024)Abdin, Aneja, Awadalla, Awadallah, Awan, Bach, Bahree, Bakhtiari, Bao, Behl, et~al.]{abdin2024phi}
Marah Abdin, Jyoti Aneja, Hany Awadalla, Ahmed Awadallah, Ammar~Ahmad Awan, Nguyen Bach, Amit Bahree, Arash Bakhtiari, Jianmin Bao, Harkirat Behl, et~al.
\newblock Phi-3 technical report: A highly capable language model locally on your phone.
\newblock \emph{arXiv preprint arXiv:2404.14219}, 2024.

\bibitem[Achiam et~al.(2023)Achiam, Adler, Agarwal, Ahmad, Akkaya, Aleman, Almeida, Altenschmidt, Altman, Anadkat, et~al.]{achiam2023gpt}
Josh Achiam, Steven Adler, Sandhini Agarwal, Lama Ahmad, Ilge Akkaya, Florencia~Leoni Aleman, Diogo Almeida, Janko Altenschmidt, Sam Altman, Shyamal Anadkat, et~al.
\newblock Gpt-4 technical report.
\newblock \emph{arXiv preprint arXiv:2303.08774}, 2023.

\bibitem[Bai et~al.(2023)Bai, Bai, Chu, Cui, Dang, Deng, Fan, Ge, Han, Huang, et~al.]{bai2023qwen}
Jinze Bai, Shuai Bai, Yunfei Chu, Zeyu Cui, Kai Dang, Xiaodong Deng, Yang Fan, Wenbin Ge, Yu~Han, Fei Huang, et~al.
\newblock Qwen technical report.
\newblock \emph{arXiv preprint arXiv:2309.16609}, 2023.

\bibitem[Baumann et~al.(2024)Baumann, Ghignone, K{\"u}hne, Bastuck, Becker, Imholz, Kr{\"a}nzlin, Lim, L{\"o}tscher, Schwarzenbach, et~al.]{forzaeth}
Nicolas Baumann, Edoardo Ghignone, Jonas K{\"u}hne, Niklas Bastuck, Jonathan Becker, Nadine Imholz, Tobias Kr{\"a}nzlin, Tian~Yi Lim, Michael L{\"o}tscher, Luca Schwarzenbach, et~al.
\newblock Forzaeth race stack—scaled autonomous head-to-head racing on fully commercial off-the-shelf hardware.
\newblock \emph{Journal of Field Robotics}, 2024.

\bibitem[Berger and Rumpe(2012)]{berger2012engineering}
Christian Berger and Bernhard Rumpe.
\newblock Engineering autonomous driving software.
\newblock \emph{Experience from the DARPA Urban Challenge}, pages 243--271, 2012.

\bibitem[Bolte et~al.(2019)Bolte, Bar, Lipinski, and Fingscheidt]{bolte2019towards}
Jan-Aike Bolte, Andreas Bar, Daniel Lipinski, and Tim Fingscheidt.
\newblock Towards corner case detection for autonomous driving.
\newblock In \emph{2019 IEEE Intelligent vehicles symposium (IV)}, pages 438--445. IEEE, 2019.

\bibitem[Caesar et~al.(2020)Caesar, Bankiti, Lang, Vora, Liong, Xu, Krishnan, Pan, Baldan, and Beijbom]{nuscenes}
Holger Caesar, Varun Bankiti, Alex~H Lang, Sourabh Vora, Venice~Erin Liong, Qiang Xu, Anush Krishnan, Yu~Pan, Giancarlo Baldan, and Oscar Beijbom.
\newblock nuscenes: A multimodal dataset for autonomous driving.
\newblock In \emph{Proceedings of the IEEE/CVF conference on computer vision and pattern recognition}, pages 11621--11631, 2020.

\bibitem[Chavan et~al.(2024)Chavan, Magazine, Kushwaha, Debbah, and Gupta]{chavan2024faster}
Arnav Chavan, Raghav Magazine, Shubham Kushwaha, M{\'e}rouane Debbah, and Deepak Gupta.
\newblock Faster and lighter llms: A survey on current challenges and way forward.
\newblock \emph{arXiv preprint arXiv:2402.01799}, 2024.

\bibitem[Chen and Ran(2019)]{iot_good1}
Jiasi Chen and Xukan Ran.
\newblock Deep learning with edge computing: A review.
\newblock \emph{Proceedings of the IEEE}, 107\penalty0 (8):\penalty0 1655--1674, 2019.
\newblock \doi{10.1109/JPROC.2019.2921977}.

\bibitem[Daniel~Han and team(2023)]{unsloth}
Michael~Han Daniel~Han and Unsloth team.
\newblock Unsloth, 2023.
\newblock URL \url{http://github.com/unslothai/unsloth}.

\bibitem[Duan et~al.(2024)Duan, Yuan, Pumacay, Wang, Ehsani, Fox, and Krishna]{duan2024manipulate}
Jiafei Duan, Wentao Yuan, Wilbert Pumacay, Yi~Ru Wang, Kiana Ehsani, Dieter Fox, and Ranjay Krishna.
\newblock Manipulate-anything: Automating real-world robots using vision-language models.
\newblock \emph{arXiv preprint arXiv:2406.18915}, 2024.

\bibitem[Fuchs et~al.(2021)Fuchs, Song, Kaufmann, Scaramuzza, and Dürr]{rl_works0}
Florian Fuchs, Yunlong Song, Elia Kaufmann, Davide Scaramuzza, and Peter Dürr.
\newblock Super-human performance in gran turismo sport using deep reinforcement learning.
\newblock \emph{IEEE Robotics and Automation Letters}, 6\penalty0 (3):\penalty0 4257--4264, 2021.
\newblock \doi{10.1109/LRA.2021.3064284}.

\bibitem[Geiger et~al.(2013)Geiger, Lenz, Stiller, and Urtasun]{geiger2013vision}
Andreas Geiger, Philip Lenz, Christoph Stiller, and Raquel Urtasun.
\newblock Vision meets robotics: The kitti dataset.
\newblock \emph{The International Journal of Robotics Research}, 32\penalty0 (11):\penalty0 1231--1237, 2013.

\bibitem[Gerganov and Contributors(2023)]{llamacpp}
Georgi Gerganov and Ope-Source Contributors.
\newblock Llama.cpp, 2023.
\newblock URL \url{https://github.com/ggerganov/llama.cpp}.

\bibitem[Guo et~al.(2025)Guo, Yang, Zhang, Song, Zhang, Xu, Zhu, Ma, Wang, Bi, et~al.]{r1}
Daya Guo, Dejian Yang, Haowei Zhang, Junxiao Song, Ruoyu Zhang, Runxin Xu, Qihao Zhu, Shirong Ma, Peiyi Wang, Xiao Bi, et~al.
\newblock Deepseek-r1: Incentivizing reasoning capability in llms via reinforcement learning.
\newblock \emph{arXiv preprint arXiv:2501.12948}, 2025.

\bibitem[Heidecker et~al.(2021)Heidecker, Breitenstein, R{\"o}sch, L{\"o}hdefink, Bieshaar, Stiller, Fingscheidt, and Sick]{heidecker2021application}
Florian Heidecker, Jasmin Breitenstein, Kevin R{\"o}sch, Jonas L{\"o}hdefink, Maarten Bieshaar, Christoph Stiller, Tim Fingscheidt, and Bernhard Sick.
\newblock An application-driven conceptualization of corner cases for perception in highly automated driving.
\newblock In \emph{2021 IEEE Intelligent Vehicles Symposium (IV)}, pages 644--651. IEEE, 2021.

\bibitem[Hu et~al.(2021)Hu, Shen, Wallis, Allen-Zhu, Li, Wang, Wang, and Chen]{lora}
Edward~J Hu, Yelong Shen, Phillip Wallis, Zeyuan Allen-Zhu, Yuanzhi Li, Shean Wang, Lu~Wang, and Weizhu Chen.
\newblock Lora: Low-rank adaptation of large language models.
\newblock \emph{arXiv preprint arXiv:2106.09685}, 2021.

\bibitem[Huang et~al.(2023)Huang, Yu, Ma, Zhong, Feng, Wang, Chen, Peng, Feng, Qin, et~al.]{huang2023surveyhallucination}
Lei Huang, Weijiang Yu, Weitao Ma, Weihong Zhong, Zhangyin Feng, Haotian Wang, Qianglong Chen, Weihua Peng, Xiaocheng Feng, Bing Qin, et~al.
\newblock A survey on hallucination in large language models: Principles, taxonomy, challenges, and open questions.
\newblock \emph{ACM Transactions on Information Systems}, 2023.

\bibitem[Hurst et~al.(2024)Hurst, Lerer, Goucher, Perelman, Ramesh, Clark, Ostrow, Welihinda, Hayes, Radford, et~al.]{hurst2024gpt4o}
Aaron Hurst, Adam Lerer, Adam~P Goucher, Adam Perelman, Aditya Ramesh, Aidan Clark, AJ~Ostrow, Akila Welihinda, Alan Hayes, Alec Radford, et~al.
\newblock Gpt-4o system card.
\newblock \emph{arXiv preprint arXiv:2410.21276}, 2024.

\bibitem[Ismail et~al.(2024)Ismail, Arbues, Cotterell, Zurbr{\"u}gg, and Alonso]{ismail2024narrate}
Seif Ismail, Antonio Arbues, Ryan Cotterell, Ren{\'e} Zurbr{\"u}gg, and Carmen~Amo Alonso.
\newblock Narrate: Versatile language architecture for optimal control in robotics.
\newblock \emph{arXiv preprint arXiv:2403.10762}, 2024.

\bibitem[Lee et~al.(2020)Lee, Hwangbo, Wellhausen, Koltun, and Hutter]{nn_heuristics1}
Joonho Lee, Jemin Hwangbo, Lorenz Wellhausen, Vladlen Koltun, and Marco Hutter.
\newblock Learning quadrupedal locomotion over challenging terrain.
\newblock \emph{Science Robotics}, 5\penalty0 (47):\penalty0 eabc5986, 2020.
\newblock \doi{10.1126/scirobotics.abc5986}.
\newblock URL \url{https://www.science.org/doi/abs/10.1126/scirobotics.abc5986}.

\bibitem[Levinson et~al.(2011)Levinson, Askeland, Becker, Dolson, Held, Kammel, Kolter, Langer, Pink, Pratt, et~al.]{levinson2011towards}
Jesse Levinson, Jake Askeland, Jan Becker, Jennifer Dolson, David Held, Soeren Kammel, J~Zico Kolter, Dirk Langer, Oliver Pink, Vaughan Pratt, et~al.
\newblock Towards fully autonomous driving: Systems and algorithms.
\newblock In \emph{2011 IEEE intelligent vehicles symposium (IV)}, pages 163--168. IEEE, 2011.

\bibitem[Lu et~al.(2024)Lu, Li, Cai, Yi, Liu, Zhang, Lane, and Xu]{slm}
Zhenyan Lu, Xiang Li, Dongqi Cai, Rongjie Yi, Fangming Liu, Xiwen Zhang, Nicholas~D Lane, and Mengwei Xu.
\newblock Small language models: Survey, measurements, and insights.
\newblock \emph{arXiv preprint arXiv:2409.15790}, 2024.

\bibitem[Ma et~al.(2023)Ma, Liang, Wang, Huang, Bastani, Jayaraman, Zhu, Fan, and Anandkumar]{ma2023eureka}
Yecheng~Jason Ma, William Liang, Guanzhi Wang, De-An Huang, Osbert Bastani, Dinesh Jayaraman, Yuke Zhu, Linxi Fan, and Anima Anandkumar.
\newblock Eureka: Human-level reward design via coding large language models.
\newblock \emph{arXiv preprint arXiv:2310.12931}, 2023.

\bibitem[Pavone(2024)]{pavone2024gtc}
Marco Pavone.
\newblock Decision making and control with llms.
\newblock Lecture presented at NVIDIA GTC 2024, 2024.
\newblock URL \url{https://www.nvidia.com/en-us/on-demand/session/gtc24-s62855/}.
\newblock Accessed: 2024-07-02.

\bibitem[Siegwart et~al.(2011)Siegwart, Nourbakhsh, and Scaramuzza]{siegwart_amr}
Roland Siegwart, Illah~R. Nourbakhsh, and Davide Scaramuzza.
\newblock \emph{Introduction to Autonomous Mobile Robots}.
\newblock The MIT Press, 2nd edition, 2011.
\newblock ISBN 0262015358.

\bibitem[Swaminathan et~al.(2024)Swaminathan, Silver, and Akilan]{iot_good0}
Tushar~Prasanna Swaminathan, Christopher Silver, and Thangarajah Akilan.
\newblock Benchmarking deep learning models on nvidia jetson nano for real-time systems: An empirical investigation.
\newblock \emph{arXiv preprint arXiv:2406.17749}, 2024.

\bibitem[Wang et~al.(2024)Wang, Mao, Zhu, Xu, Lyu, Li, Chen, Zhang, Chen, Xue, et~al.]{wang2024embodiedscan}
Tai Wang, Xiaohan Mao, Chenming Zhu, Runsen Xu, Ruiyuan Lyu, Peisen Li, Xiao Chen, Wenwei Zhang, Kai Chen, Tianfan Xue, et~al.
\newblock Embodiedscan: A holistic multi-modal 3d perception suite towards embodied ai.
\newblock In \emph{Proceedings of the IEEE/CVF Conference on Computer Vision and Pattern Recognition}, pages 19757--19767, 2024.

\bibitem[Wen et~al.(2023)Wen, Fu, Li, Cai, Ma, Cai, Dou, Shi, He, and Qiao]{wen2023dilu}
Licheng Wen, Daocheng Fu, Xin Li, Xinyu Cai, Tao Ma, Pinlong Cai, Min Dou, Botian Shi, Liang He, and Yu~Qiao.
\newblock Dilu: A knowledge-driven approach to autonomous driving with large language models.
\newblock \emph{arXiv preprint arXiv:2309.16292}, 2023.

\bibitem[Wurman et~al.(2022)Wurman, Barrett, Kawamoto, MacGlashan, Subramanian, Walsh, Capobianco, Devlic, Eckert, Fuchs, et~al.]{rl_works1}
Peter~R Wurman, Samuel Barrett, Kenta Kawamoto, James MacGlashan, Kaushik Subramanian, Thomas~J Walsh, Roberto Capobianco, Alisa Devlic, Franziska Eckert, Florian Fuchs, et~al.
\newblock Outracing champion gran turismo drivers with deep reinforcement learning.
\newblock \emph{Nature}, 602\penalty0 (7896):\penalty0 223--228, 2022.

\bibitem[Yu et~al.(2023)Yu, Gileadi, Fu, Kirmani, Lee, Arenas, Chiang, Erez, Hasenclever, Humplik, et~al.]{l2r}
Wenhao Yu, Nimrod Gileadi, Chuyuan Fu, Sean Kirmani, Kuang-Huei Lee, Montse~Gonzalez Arenas, Hao-Tien~Lewis Chiang, Tom Erez, Leonard Hasenclever, Jan Humplik, et~al.
\newblock Language to rewards for robotic skill synthesis.
\newblock \emph{arXiv preprint arXiv:2306.08647}, 2023.

\bibitem[Zarrouki et~al.(2021)Zarrouki, Kl{\"o}s, Heppner, Schwan, Ritschel, and Vo{\ss}winkel]{nn_heuristics0}
Baha Zarrouki, Verena Kl{\"o}s, Nikolas Heppner, Simon Schwan, Robert Ritschel, and Rick Vo{\ss}winkel.
\newblock Weights-varying mpc for autonomous vehicle guidance: a deep reinforcement learning approach.
\newblock In \emph{2021 European Control Conference (ECC)}, pages 119--125. IEEE, 2021.

\end{thebibliography}
